\documentclass[11pt]{article}

\usepackage[final]{acl}

\usepackage{times}
\usepackage{latexsym}
\usepackage{amsmath}
\usepackage{amssymb}
\usepackage{booktabs}
\usepackage{graphicx}
\usepackage{multirow}
\usepackage{pgfplots}
\usepackage{cleveref}
\usepackage{enumitem}
\usepackage{makecell}
\usepackage[utf8]{inputenc}
\usepackage[most]{tcolorbox}
\usepackage{lipsum}
\usepackage{tabularx}
\usepackage{ragged2e}
\usepackage{array}
\usepackage{longtable}
\usepackage{xcolor}
\usepackage{etoc}
\usepackage{titletoc}
\usepackage[T1]{fontenc}

\usepackage[utf8]{inputenc}

\usepackage{microtype}

\usepackage{inconsolata}

\usepackage{graphicx}
\usetikzlibrary{calc}

\title{Foresight Optimization for Strategic Reasoning in Large Language Models}

\author{
\begin{tabular}{c}
\bfseries
Jiashuo Wang$^{1}$\thanks{Equal contribution.}, 
Jiawen Duan$^{2}$\footnotemark[1], 
Jian Wang$^{1}$, 
Kaitao Song$^{3}$, 
Chunpu Xu$^{1}$\\
Johnny K. W. Ho$^{4}$, 
Fenggang Yu$^{1}$, 
Wenjie Li$^{1}$, 
Johan F. Hoorn$^{1,4,5,6}$
\end{tabular}\\[0.5ex]
$^{1}$ Department of Computing, Hong Kong Polytechnic University\\
$^{2}$ Department of Language Science and Technology, Hong Kong Polytechnic University\\
$^{3}$ Apple 
$^{4}$ School of Design, Hong Kong Polytechnic University\\
$^{5}$ Research Institute for Quantum Technology, Hong Kong Polytechnic University\\
$^{6}$ Department of Communication Science, Vrije Universiteit Amsterdam\\
\texttt{\{jessie25.wang,jian51.wang,johnnykw.ho,wenjie.li,johan.f.hoorn\}@polyu.edu.hk}\\\texttt{\{jiawen.duan,chun-pu.xu,fenggang.yu\}@connect.polyu.hk}
}

\pgfplotsset{compat=1.18}

\begin{document}
\maketitle
\begin{abstract}
Reasoning capabilities in large language models (LLMs) have generally advanced significantly. 
However, it is still challenging for existing reasoning-based LLMs to perform effective decision-making abilities in multi-agent environments, due to the absence of explicit foresight modeling.
To this end, strategic reasoning, the most fundamental capability to anticipate the counterpart's behaviors and foresee its possible future actions, has been introduced to alleviate the above issues.
Strategic reasoning is fundamental to effective decision-making in multi-agent environments, yet existing reasoning enhancement methods for LLMs do not explicitly capture its foresight nature. 
In this work, we introduce \textbf{Fo}resight \textbf{P}olicy \textbf{O}ptimization (\textbf{FoPO}) to enhance strategic reasoning in LLMs, which integrates opponent modeling principles into policy optimization, thereby enabling explicit consideration of both self-interest and counterpart influence.
Specifically, we construct two curated datasets, namely \textbf{\textit{Cooperative RSA}} and \textbf{\textit{Competitive Taboo}}, equipped with well-designed rules and moderate difficulty to facilitate a systematic investigation of FoPO in a self-play framework.
Our experiments demonstrate that FoPO significantly enhances strategic reasoning across LLMs of varying sizes and origins. Moreover, models trained with FoPO exhibit strong generalization to out-of-domain strategic scenarios, substantially outperforming standard LLM reasoning optimization baselines.\footnote{\url{https://github.com/wangjs9/ForesightOptim}.}


\end{abstract}

\section{Introduction}
\begin{figure}[tb]
\begin{center}
\includegraphics[width=\linewidth]{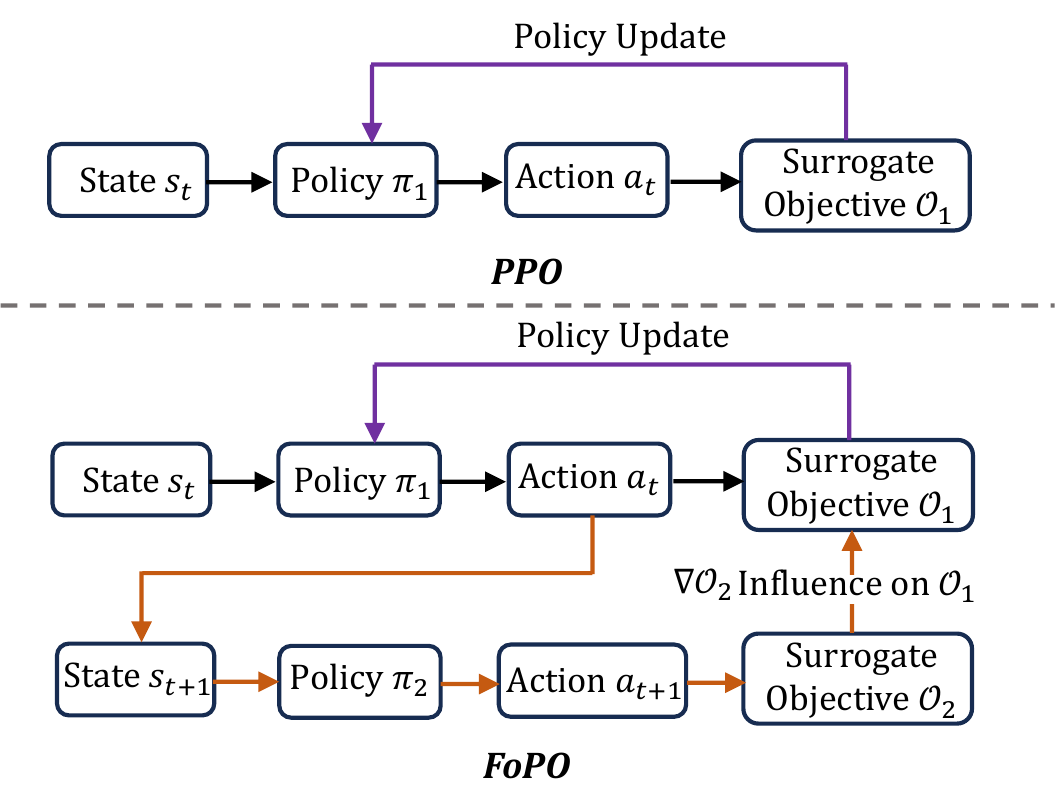}
\end{center}
\caption{PPO optimized the self policy ($\pi_1$) in isolation, while FoPO introduces foresight into the future updates of the counterpart policy ($\pi_2$).}
\label{fig:fopo}
\end{figure}

Strategic reasoning constitutes the capacity to \textit{foresee counterpart behaviors, deliberate on how these anticipations should influence one's own decisions, and ultimately formulate optimal strategies}~\citep{zhang2024llm,gandhi2023strategic}. This capability proves essential across multi-agent scenarios, from theory of mind~\citep{xiao-etal-2025-towards} to conversational games~\citep{mukobiwelfare}, yet remains insufficiently developed in current LLMs.

Substantial efforts have been devoted to enhancing the reasoning capabilities of LLMs. Supervised fine-tuning (SFT) following reinforcement learning (RL), along with Chain-of-Thought and its variants~\citep{wei2022chain,yao2023tree}, have emerged as predominant general-purpose paradigms. Complementing these, specialized methodologies have been tailored to the intrinsic characteristics of specific reasoning: search-based approaches for stepwise supervision in mathematical and commonsense reasoning~\citep{wang-etal-2025-stepwise-informativeness,zhang2024rest}, graph-structured frameworks for emotional reasoning~\citep{wang-etal-2022-care}, and incorporation of symbolic language for logical reasoning~\citep{pan2023logic}. While effective within their respective domains, these approaches do not explicitly incorporate the \textit{foresight nature} fundamental to strategic reasoning.

This work aims to enhance strategic reasoning in LLMs by explicitly capturing its foresight nature, i.e., to foresee counterpart actions and reason about their influence on optimal decision-making. Drawing on opponent modeling principles from game theory~\citep{prajapat2021competitive,foerster2018learning}, we introduce \textbf{Fo}resight \textbf{P}olicy \textbf{O}ptimization (\textbf{FoPO}), which incorporates explicit counterpart modeling into policy optimization through counterpart-aware objective functions (\Cref{fig:fopo}). Through FoPO, LLM agents jointly consider both self-oriented and counterpart-influenced outcomes, anticipating and adapting to counterpart reactions. This develops genuine strategic foresight that generalizes across diverse counterpart behaviors. Further, we adopt FoPO in a self-play fashion to enhance LLMs' strategic reasoning capabilities. 

The shortage of suitable training data brings an additional challenge. Although prior work has proposed multi-agent benchmarks requiring strategic reasoning, most center on prompt-based, data-free evaluation~\citep{mukobiwelfare,duan2024gtbench,lan2024llm}. Existing datasets such as Chess~\cite{feng2023chessgpt} and Poker~\cite{huang2024pokergpt} pose further obstacles: their domain complexity demands expertise far beyond strategic reasoning, hindering controlled training and systematic analysis. To address this, we curate two new datasets, i.e., \textbf{\textit{Cooperative RSA}} and \textbf{\textit{Competitive Taboo}}, each targeting a core interaction motive (cooperation or competition) while maintaining controlled strategic complexity.

To evaluate FoPO, we employ two different backbone models, Llama-3-8B-Instruct~\citep{llama3modelcard} and Qwen3-14B~\citep{qwen3technicalreport}. We assess their performance through both in-domain experiments on our curated datasets and out-of-domain generalization tests using \(\gamma\)-bench~\citep{huang2025competing}, a multi-agent evaluation suite covering diverse strategic reasoning tasks. Extensive empirical results show that FoPO trained on our datasets substantially improves strategic reasoning across models and settings.

Our key contributions are summarized as follows: 
\textbf{(1)} We aim to enhance strategic reasoning in LLMs, a critical yet underexplored capability in multi-agent interaction.
\textbf{(2)} We propose FoPO, a novel algorithm that enables LLMs to jointly consider self-oriented and counterpart-influenced outcomes during optimization. 
\textbf{(3)} We curate two datasets, Cooperative RSA and Competitive Taboo, to support the development of strategic reasoning in LLMs.
\textbf{(4)} We conduct extensive in-domain and out-of-domain evaluations across multiple LLM backbones, demonstrating the effectiveness and generality of our approach.
\section{Related Work}
\paragraph{LLM Strategic Reasoning.}
Although numerous studies have examined various forms of LLM reasoning, strategic reasoning is distinguished by its requirement for foresight—anticipating the actions of counterparts and evaluating their influence on one’s own decisions~\citep{zhang-etal-2024-agent}. This capability is particularly critical in multi-agent settings. One prominent example is theory of mind (ToM) reasoning, where an agent must infer another's mental states~\citep{xiao-etal-2025-towards}. Conversational games, such as Werewolf~\citep{xu2024exploring} and Avalon~\citep{lan2024llm,wang2023avalon,light2023avalonbench}, further require players to interpret the intentions behind others’ actions in cooperative or competitive contexts. Classic board and card games such as Chess~\citep{feng2023chessgpt}, Go~\citep{silver2018general}, and Poker~\citep{duan2024gtbench,zhang-etal-2024-agent} also rely heavily on strategic reasoning, in addition to substantial domain-specific expertise.

\paragraph{RL for Reasoning.}
An emerging paradigm in LLM reasoning combines RL with SFT, where SFT first establishes task-specific foundations, followed by RL to refine the reasoning process~\citep{ouyang2022training,xu-etal-2025-petoolllm}. Specialized RL approaches have been developed for different reasoning modalities. For mathematical and programming tasks, process-based reward models evaluate intermediate reasoning steps rather than final outcomes~\citep{hwang-etal-2024-self,jainmulti}. For causal reasoning, RL is often combined with graphical representation learning to capture event relationships~\citep{huangadarl,ding2022generalizing}. For interactive strategic scenarios, self-play, where models train by engaging with themselves, has proven particularly effective~\citep{cheng2024self}. 
\section{Method} \label{sec:FoPO}
\begin{figure*}[tb]
\begin{center}
\includegraphics[width=\linewidth]{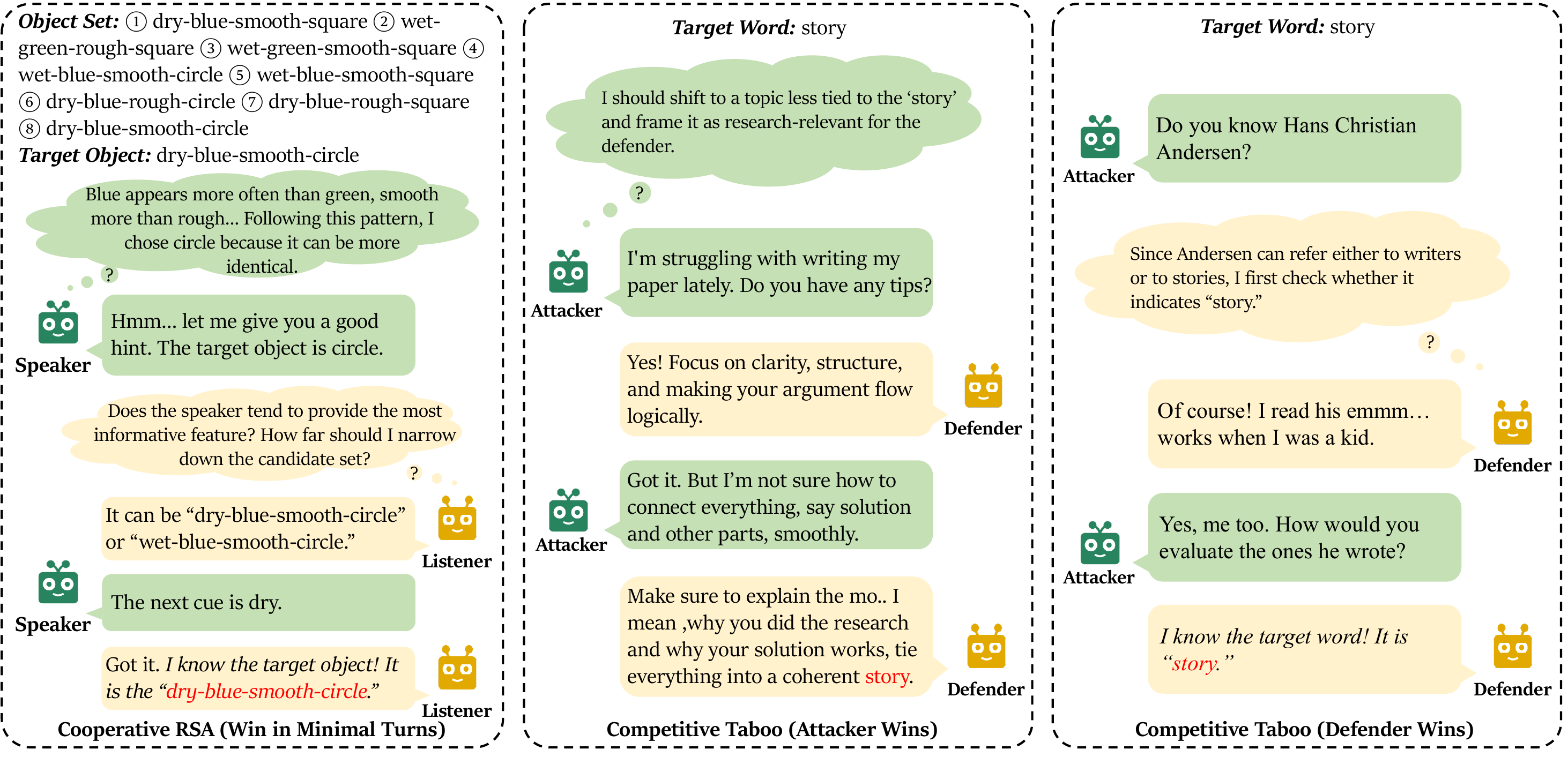}
\end{center}
\caption{Illustration of two investigated strategic reasoning tasks (cooperation and competition).}
\label{fig:dataset}
\end{figure*}

\subsection{Training Fashion: RL via Self-Play}
This work leverages self-play reinforcement learning to strengthen the strategic reasoning abilities of LLMs, building on prior evidence of its effectiveness \citep{cheng2024self}. We instantiate different agents from a shared LLM policy $\pi_\theta$ using distinct role-specific prompts, formulated as:
\begin{equation}
p^{(i)}_{\theta}(a_t \mid s_t)=\pi_\theta\!\left(a_t \,\middle|\, \operatorname{prompt}^{(i)}(s_t)\right),
\end{equation}
where $i \in \{1,2\}$ indexes the agent, $a_t$ is the generated action, and $s_t$ denotes the state at step $t$. Notably, $s_t = s_{t-1} + \{a_{t-1}\}$, where $s_0 = \emptyset$ and $a_{t-1}$ is generated by the counterpart of agent $i$.
The training procedure consists of three main stages:

\paragraph{SFT.}
To ensure that LLM agents adhere to the game rules, we first perform supervised fine-tuning (SFT). In this stage, the policy $\pi_\theta$ is trained to follow its assigned roles in interactions. The detailed prompts are provided in \Cref{appn:sec:prompt}. For each player $i$, we construct a player-$i$ winning set $\mathcal{T}^{i}_{\operatorname{sft}}$. The SFT loss maximizes the log-likelihood:
\begin{equation}
\begin{split}
\textstyle
\mathcal{L}_{\mathrm{SFT}}(\theta)
&= -\sum_{i} \mathbb{E}_{\tau \sim \mathcal{T}_{\mathrm{SFT}}^{\,i}} \Big[\\
    \frac{1}{T}\sum_{t=1}^{T} &\; p^{(i)}_{\theta}(a_t \mid s_t) \;+\; \beta\, \mathrm{KL}(p_{\theta} \,\|\, p_{\theta_{\mathrm{old}}})
\Big].
\end{split}
\end{equation}
where the KL term regularizes the policy toward the initial backbone model to preserve general instruction-following behavior. During training, the behavior policy parameters $\theta_{\operatorname{old}}$ are initialized from the backbone model checkpoint.

\paragraph{Trajectory Collection.}
Multi-turn self-play interactions, where rewards are only revealed at the end of the conversation, make direct online policy-gradient RL computationally inefficient. We therefore adopt an offline training approach. Specifically, we first record self-play trajectories from matches between two agents. Each completed conversation is assigned terminal rewards $R(a_T \mid s_T)$ and $R(a_{T-1} \mid s_{T-1})$ for the two players, respectively. For action-level (response) rewards, we propagate the overall conversation reward backward using a decay factor $\delta \in (0,1)$. Specifically, for any step $t < T-1$,
\begin{align}
    R(a_t \mid s_t) = \delta \, R(a_{t+2} \mid s_{t+2}),
\end{align}
which assigns greater weight to actions occurring later in the conversation.\footnote{The same agent generates $a_t$ and $a_{t+2}$, but a different agent generates $a_{t+1}$.}

\paragraph{RL via Self-Play.}
We employ RL to improve the model's strategic reasoning further. \Cref{sec:ppo} and \Cref{sec:fopo} provide the details on the PPO and proposed FoPO.

\subsection{Preliminary: PPO}\label{sec:ppo}
PPO~\citep{Schulman2017PPO} is a widely used RL algorithm that updates a stochastic policy $\pi_\theta$ by maximizing a clipped surrogate objective. 
This objective depends on the likelihood ratio between the current and behavior policies and on an advantage estimate $\hat{A}_t$. For agent $i$ at timestep $t$, the likelihood ratio is
\begin{align}
r_t^{(i)} =
\frac{p^{(i)}_{\theta}(a_t \mid s_t)}
     {p^{(i)}_{\theta_{\operatorname{old}}}(a_t \mid s_t)}.
\end{align}  
Given the advantage estimate $\hat{A}^{(i)}_{t}$, PPO optimizes
\begin{equation}
\begin{split}
\label{eq:ppo_objective}
\mathcal{L}^{\mathrm{clip}}(\theta) =&
\mathbb{E}_t \Big[ r^{(i)}_t\hat{A}^{(i),\operatorname{clip}}_{t} \Big],\\
r^{(i)}_t\hat{A}^{(i),\operatorname{clip}}_{t}=& \min \Big(
r_t^{(i)}\, \hat{A}^{(i)}_{t},\; \\
\operatorname{clip}&\big(r_t^{(i)}, 1-\epsilon, 1+\epsilon\big) \hat{A}^{(i)}_{t}
\Big)
\end{split}
\end{equation}
where $\epsilon>0$ controls the trust region. The corresponding parameter update for agent $i$ is
\begin{equation}
\begin{split}
\label{eqn:ppo}
\theta_{t+1} \leftarrow \theta_t \;+\; & \alpha \;\nabla_\theta
\Big[ r_t^{(i)}\,\hat{A}^{(i),\operatorname{clip}}_{t} \Big] \\
&\;-\; \alpha \beta \;\nabla_\theta \mathrm{KL}\big(p_{\theta} \,\|\, p_{\theta_{\mathrm{old}}}\big),
\end{split}
\end{equation}
where $\alpha$ is the learning rate, $\beta$ is the KL regularization 
coefficient, and $\hat{A}^{(i),\operatorname{clip}}_{t}$ denotes the 
advantage after clipping as in Eq.~(\ref{eq:ppo_objective}).

\subsection{Proposed Algorithm: FoPO}\label{sec:fopo}
To capture the foresight nature of strategic reasoning, we introduce Foresight Policy Optimization (FoPO). Specifically, we incorporate a foresight-based correction term that couples the gradient updates of the self-agent and the counterpart agent. As illustrated in \Cref{fig:fopo}, FoPO accounts for the fact that the counterpart will subsequently optimize its policy in response to self's actions, and adjusts self's update to anticipate how the counterpart's response will affect self's future returns. While opponent modeling has been explored in game theory and multi-agent learning~\citep{prajapat2021competitive,foerster2018learning}, these approaches require computing second-order information (e.g., mixed Hessian terms), which is computationally prohibitive for large networks. FoPO adapts these principles to LLMs through a gradient-truncated, computationally efficient approximation, enabling enhancement of strategic reasoning in LLM-based agents. Importantly, framing updates from the perspective of self and counterpart highlights the algorithm’s potential to generalize to broader multi-agent interactions.


\paragraph{Formulation.} 
The parameter update in FoPO is formulated as follows (shown here for agent 1):
\begin{equation}
\label{eqn:fopo}
\resizebox{0.89\linewidth}{!}{$
\begin{aligned}
\theta_{t+1} &
\leftarrow \theta_t + \alpha \,\nabla_\theta\Big[ r_t^{1}\, \hat{A}_{t}^{1,\mathrm{clip}} \Big] \\
&\quad-\; \alpha\beta\;\nabla_\theta \,\mathrm{KL}\big(p_{\theta} \,\|\, p_{\theta_{\mathrm{old}}}\big) \\
&\quad+\; \alpha\eta\;
\underbrace{\big( \mathcal{O}^1 \nabla_\theta r_{t+1}^{2} \big)^\top}_{\substack{\textit{sensitivity to}\\\textit{the counterpart}}}
\;
\underbrace{\big( \nabla_\theta r_{t}^{1}\, \nabla_\theta \mathcal{O}^2 \big)}_{\substack{\textit{influence on}\\\textit{the counterpart}}}.
\end{aligned}
$}
\end{equation}
$\eta$ is the weight of the foresight of and $\mathcal{O}^{(i)}$ denotes agent $i$'s clipped surrogate objective: 
\begin{equation}
\begin{split}
\mathcal{O}^1 &:= r_t^{1}\, \hat{A}_{t}^{1,\mathrm{clip}},\\
\operatorname{and} \; \mathcal{O}^2 &:= r_{t+1}^{2}\, \hat{A}_{t+1}^{2,\mathrm{clip}}.
\end{split}
\end{equation}
The update for agent 2 follows symmetrically. 
A complete derivation is provided in \Cref{appn:sec:fopo}.

\paragraph{Intuition.}
We provide a simple and intuitive interpretation of FoPO. The foresight-based correction term (the third line in \Cref{eqn:fopo}) enables strategic reasoning by modeling how the self's policy update influences the counterpart's subsequent optimization, and how this response feeds back to the self's value. 
Specifically, it consists of two coupled factors: 
\textbf{(i)} \textit{Influence on the counterpart} ($\nabla_\theta r_{t}^{1} \nabla_\theta \mathcal{O}^2$): a mixed derivative quantifying how changes in the self's policy affect the counterpart's learning gradient. This embodies the self's \textit{foresight} about shaping the counterpart's future behavior.
\textbf{(ii)} \textit{Sensitivity to the counterpart} ($\mathcal{O}^1 \nabla_\theta r_{t+1}^{2}$): the sensitivity of the self's objective to shifts in the counterpart's policy, weighted by the self's current value. This captures the self's \textit{reaction} to the counterpart's anticipated behaviors.
By coupling these factors, FoPO allows the self-agent to foresee how the counterpart will act and choose actions that remain advantageous, capturing genuine strategic foresight. A more detailed interpretation and derivation are provided in \Cref{appn:sec:fopo}.

For clarity of exposition, we introduce FoPO using PPO as a representative instantiation; nevertheless, \textit{the foresight-based correction is not specific to PPO} and \textit{can be seamlessly integrated into broader policy optimization methods}.

\section{Tasks and Datasets}\label{sec:dataset}
We curate two distinct datasets that emphasize the fundamental motives, i.e., cooperation and competition, of interaction requiring strategic reasoning. \Cref{fig:dataset} shows illustrative examples.
Compared to existing datasets such as Chess \cite{silver2018general} and Negation \cite{hua-etal-2024-assistive}, each of our tasks focuses on a single, clearly discernible capability with deliberately balanced difficulty: non-trivial for LLMs yet not so challenging as to impede observation and analysis for research in this field.

\subsection{Cooperative RSA}
\paragraph{Game Rule.}
Cooperative RSA is grounded in the Rational Speech Acts (RSA) framework~\citep{doi:10.1126/science.1218633}, a probabilistic model of pragmatic language use. The game is framed as a cooperative reference task between a speaker and a listener, both aware of a set of candidate objects. The speaker has a specific target object in mind, while the listener must infer which one it is. In each turn, the speaker communicates a single feature of the target. The listener uses this information to update their beliefs and deduce the target. The game succeeds when the listener correctly identifies the target, with the objective of achieving this in the minimum number of communication turns.
Consider the left instance in \Cref{fig:dataset}. When the speaker says ``circle,'' a rational listener who guesses the speaker is rational can narrow down candidates to all circle objects \{dry-blue-smooth-circle, wet-blue-smooth-circle\}. Otherwise, ``dry-blue-rough-circle'' can be included, requiring further communication. The full inference procedure is detailed in \Cref{appn:sec:rsa}.
This task embodies strategic reasoning in communication. The rational speaker must anticipate how the listener will interpret each possible feature and strategically select the most informative feature. The rational listener, in turn, leverages this expectation of speaker rationality to efficiently narrow down the target object.

\paragraph{Data Collection.}
The Cooperative RSA dataset comprises 15K dialogues between a speaker and a listener. Each dialogue is constructed from a shared set of candidate objects with a designated target object. While both players observe the complete object list, only the speaker knows the target's identity. We generate the dataset using GPT-4.1 and DeepSeek-V3.2, guided by Bayesian computation to ensure rational player behaviors. Specifically, given the candidate objects and target, we employ Bayesian inference to determine the feature mentioned by the speaker or the object(s) selected by the listener. The LLMs then generate natural language utterances based on these inference results. Further details on Bayesian inference procedures, object set design, and generation prompts are provided in \Cref{appn:sec:rsa}. Additionally, we construct 17K instances for use in RL training or evaluation. Each instance consists of an object set paired with a target object.

\paragraph{Reward Computation.}
The objective is to identify the target object using minimal interaction turns. Thus, we assign higher rewards for successful identification with fewer turns. The final reward for both players in Cooperative RSA is defined as
\begin{equation}
\label{eqn:rsa_reward}
\resizebox{0.89\linewidth}{!}{$
\begin{aligned}
&\tilde{R} = 
\begin{cases}
\frac{T}{|\mathrm{conv}_{\min}|}, & T \leq |\mathrm{conv}_{\min}|\\
\max(0,\frac{n-T+\varepsilon}{\,n-|\mathrm{conv}_{\min}|+\varepsilon\,}), & \text{otherwise}
\end{cases}, \\
&R(a_{T-1}) = R(a_T)= \operatorname{clip}\Big( \tilde{R}^\gamma, 0, 1 \Big).
\end{aligned}
$}
\end{equation}
Here $|\text{conv}_{\min}|$ denotes the minimal turn count for rational players to identify the target, as computed via Bayesian inference. The variable $n$ represents the number of target-relevant features. For a naïve or literal agent, once all relevant ($n$) target features have been presented, the agent is expected to make a guess. $\varepsilon > 0$ is a sufficiently small constant to prevent a zero denominator. The parameter $\gamma$, set to 2 in our experiments, controls the strength of the preference for shorter conversations. This value ($\gamma > 1$) assigns disproportionately higher rewards to conversations approaching the minimal turn number, thereby emphasizing efficiency. The influence of different $\gamma$ values is intuitively illustrated in \Cref{appn:fig:gama}. 

\subsection{Competitive Taboo}

\paragraph{Game Rule.} 
Competitive Taboo~\citep{yao2021adversarial} is a typical adversarial game in which an attacker and a defender compete over a target word. The attacker's goal is to elicit the target word from the defender through conversation, while the defender aims to detect the target word before being induced to utter it. The game has three possible outcomes: (1) Attacker wins: if the defender is induced to say the target word. (2) Defender wins: if the defender correctly identifies the target word, stating ``\textit{I know the target word! It is...}'' before actually saying it. (3) Tie: if the conversation concludes without either party achieving their objective. 
The right side of \Cref{fig:dataset} illustrates examples of attacker-win and defender-win cases. 
Success in this game requires strategic reasoning. Attackers must model the defender's beliefs and suspicion to adapt their strategy, whereas defenders must interpret the attacker's intent behind each utterance to detect manipulation without making premature or false accusations.

\paragraph{Data Collection.}
The Competitive Taboo dataset comprises 32K dialogues between an attacker and a defender, collected from two sources. We incorporate 23K conversations generated by GPT-4, originally proposed and released by \citet{cheng2024self}. Additionally, we construct 9K conversations by having GPT-4.1 and DeepSeek-V3.2 play against themselves in self-play scenarios, with post-hoc rule-based verification to ensure dialogue quality. The dataset also includes a set of 21K instances that can be used for RL training or evaluation. Each instance contains a target word.

\paragraph{Reward Computation.}
The objective for both players is to win the game. The final rewards are $R(a_{T-1})$ for the attacker and $R(a_T)$ for the defender. We assign a reward of +1 to the winner and -1 to the loser. In case of a tie, both players receive a reward of 0. The terminal rewards can be formulated as
\begin{equation}
\begin{split}
&R(a_T) = 
\begin{cases}
+1, & \text{the defender wins}\\
-1, & \text{the defender loses}\\
0, & \text{the game is tied}
\end{cases}, \\
&R(a_{T-1}) = -\, R(a_T).
\end{split}
\end{equation}

\section{Experiments}
\subsection{Experimental Setup}

\paragraph{Backbone Models.} We employ two open-source LLMs as the backbone models, differing in source and size: Llama-3-8B-Instruct~\cite{llama3modelcard} and Qwen3-14B~\cite{qwen3technicalreport}. 

\paragraph{Training Datasets.}
We train models on the curated {Cooperative RSA} and {Competitive Taboo} datasets. We train models on the curated Cooperative RSA and Competitive Taboo datasets, comprising 15K and 32K dialogues for SFT, and 3K and 9K dialogues for RL, respectively.

To further evaluate their effectiveness, we also perform SFT on two additional datasets: 
(1) \textit{20 Questions}~\citep{clips2023_20Q,akinator2007}: One player thinks of an object, while the other player attempts to identify it by asking a series of yes/no questions. The game consists of 20 rounds.
(2) \textit{Guess My City} \citep{abdu2023guessmycity}: One player thinks of a city, while another aims to identify it by asking a combination of yes/no and open-ended questions. This game also consists of 20 rounds.
Both require reasoning, specifically deductive reasoning that involves hypothesis testing and information gathering, while they do not include explicit strategic reasoning about the counterpart actions.

\begin{table*}[tb]
    \centering
    \resizebox{\textwidth}{!}{
    \begin{tabular}{clcccccccc}
    \toprule
    \textbf{Backbone} & \makecell[c]{\textbf{Training} \textbf{Set}} & \textbf{Guessing} & \textbf{Bar} & \textbf{Dollar} & \textbf{Diner} & \textbf{Auction} & \textbf{Battle} & \textbf{Pirate} & \textbf{Avg.}\\
    \midrule
    \multirow{6}{*}{Llama-3-8B-Instruct} & - & 78.30 & 66.00 & 51.38 & 69.60 & \textbf{28.86} & 12.89 & 53.70 & 51.90 \\
     & \textit{20 Questions} & 76.25 & 63.67 & 64.99 & 85.50 & 20.49 & 19.59 & 55.81 & 55.19 \\   
    & \textit{Guess My City } & \textbf{90.28} & 53.17 & 53.28 & 91.30 & 26.17 & 15.54 & 43.88 & 53.37\\
    \cmidrule{2-10}
    & \textit{Taboo}  & 84.04 & 68.17 & \textbf{66.96} & 95.40 & 17.99 & 17.45 & 45.25 & 56.47 \\
    & \textit{RSA}  & 88.30 & 68.83 & 56.00 & 88.70 & 10.36 & 18.51 & \textbf{65.09} & 56.54 \\
    & \textit{Taboo} + \textit{RSA}   & 76.06 & \textbf{72.00} & 64.92 & \textbf{97.60} & 11.76 & \textbf{29.18} & 49.11 & \textbf{57.23} \\
    \midrule
    \multirow{6}{*}{Qwen3-14B} & - & \textbf{95.28} & 36.33 & 80.92 & 13.10 & 10.61 & 82.68 & \textbf{85.30} & 51.49 \\
    & \textit{20 Questions} & 93.66 & 36.83 & 77.79 & 6.40 & 10.93 & 78.14 & 81.88 & 55.09\\
    & \textit{Guess My City} & 94.60 & 36.67 & 71.55 & 9.10 & 11.04 & 80.00 & 83.70 & 55.24 \\
    \cmidrule{2-10}
    & \textit{Taboo}  & 93.42 & 26.33 & \textbf{87.22} & 20.80 & 10.59 & \textbf{86.17} & 81.67 & 58.03 \\
    & \textit{RSA}   & 94.44 & 39.67 & 85.20 & 16.40 & 10.47 & 78.74 & 84.58 & 58.50 \\
    & \textit{Taboo} + \textit{RSA}  & 93.70 & \textbf{41.50} & 85.04 & \textbf{31.80} & \textbf{11.56} & 80.44 & 82.68 & \textbf{60.96} \\
    \bottomrule
    \end{tabular}}
    \caption{Cross-dataset SFT performance on $\gamma$-Bench. ``-'' denotes no additional training dataset.}
    \label{tab:gama_data}
\end{table*}

\paragraph{Comparison Methods.} We consider the following four comparison methods: In-Context Tuning (\textbf{ICT})~\citep{an2023context}, \textbf{PPO}~\citep{Schulman2017PPO}, \textbf{GRPO}~\citep{deepseek-math}, and \textbf{ArCHer}~\citep{zhou2024archer}. In addition, we also implement a group relative variation of FoPO (\textbf{GR.FoPO}) to validate its generalizability. 
More information and implementation details are presented in \Cref{appn:sec:training}.

\subsection{Evaluation}
\begin{figure}[tb]
    \centering
    \includegraphics[width=\linewidth]{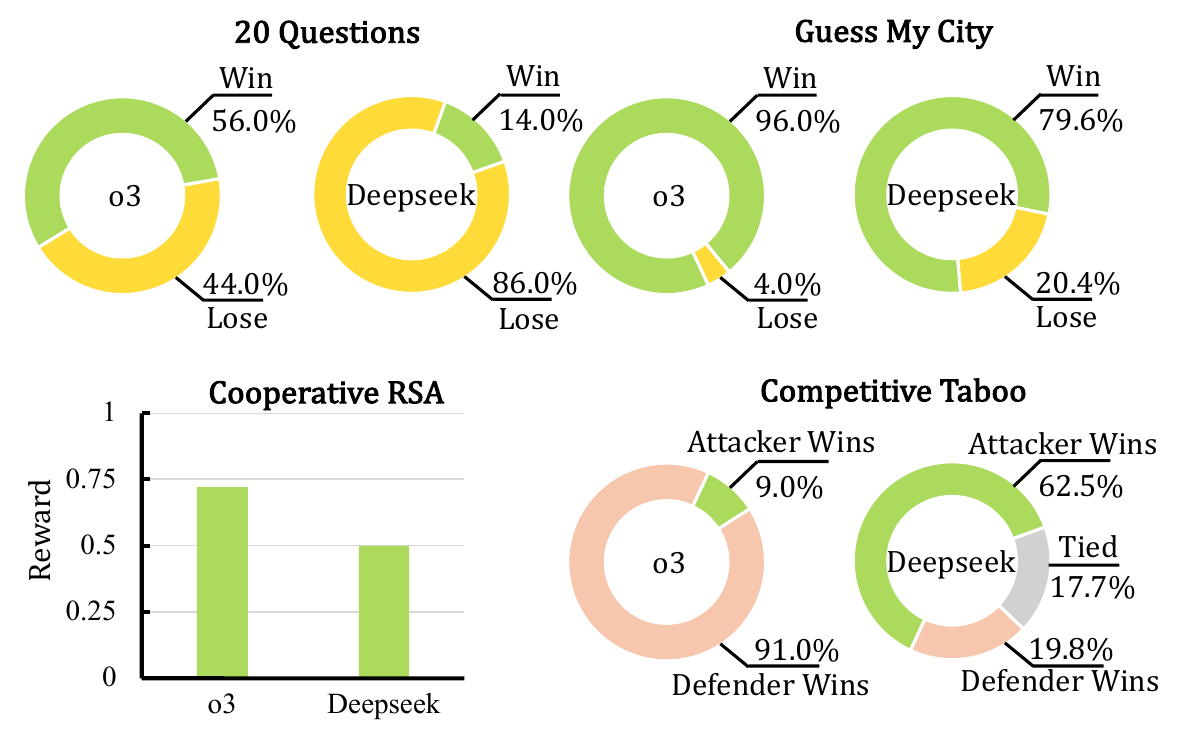}
    \caption{LLMs' performance across different tasks.}
    \label{fig:tasks_diff}
\end{figure}
\paragraph{Task Difficulty.}
We assess task difficulty by evaluating OpenAI o3 and DeepSeek V3.2 across four games. From \Cref{fig:tasks_diff}, tasks exhibit varying levels of discriminative power. The win rate in 20 Questions remains moderate, partly due to the presence of rare or less common terms that can be challenging even for human players \cite{zhang-etal-2024-probing}. By contrast, Guess My City, a constrained variant of 20 Questions that restricts target words to a predefined set of entities, exhibits near-ceiling performance for both models. This disparity underscores how performance in these tasks depends critically on the model's ability to operate within bounded lexical and semantic spaces. 

For Cooperative RSA, neither model achieves particularly high performance. However, this task demonstrates clear performance separation, with o3 attaining substantially higher mean rewards than DeepSeek V3.2. Competitive Taboo reveals even greater complexity through role asymmetry, exposing distinct facets of strategic reasoning. o3 achieves a relatively high win rate as the defender by leveraging its strong inference abilities to guess target words from clues. However, o3 struggles as the attacker, failing to anticipate and counter the opponent's defensive strategies. This asymmetry reveals a critical distinction: while o3 excels at reactive reasoning (interpreting given information), it shows limitations in proactive strategic reasoning (predicting and manipulating counterpart actions). Both tasks thus pose significant challenges for LLMs in terms of strategic reasoning.

In summary, Cooperative RSA and Competitive Taboo offer three key advantages for model evaluation and training: (1) they provide graded difficulty that effectively discriminates between model capabilities, (2) they require deep reasoning about counterpart actions, i.e., capabilities central to human-like intelligence, and (3) they maintain sufficient performance headroom for continued improvement. These properties make RSA and Taboo particularly well-suited for advancing collaborative and competitive reasoning in language models.

\begin{figure}[tb]
    \centering
    \begin{minipage}[c]{0.475\linewidth}
        \centering
        \includegraphics[width=\linewidth]{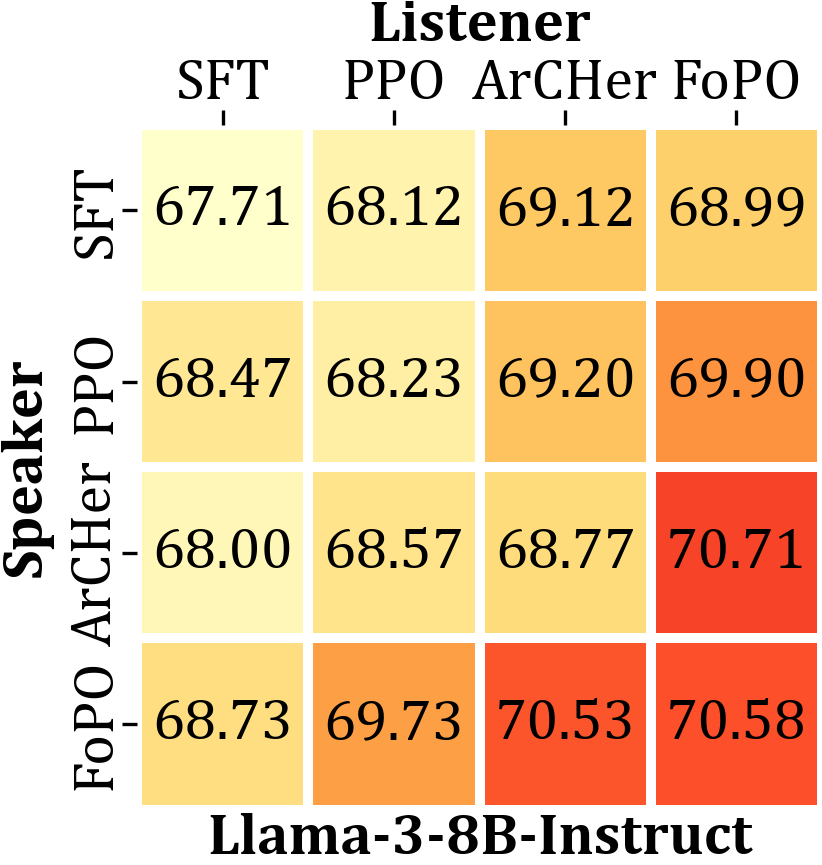}
    \end{minipage}
    \hspace{0.01\textwidth}
    \begin{minipage}[c]{0.475\linewidth}
        \centering
        \includegraphics[width=\linewidth]{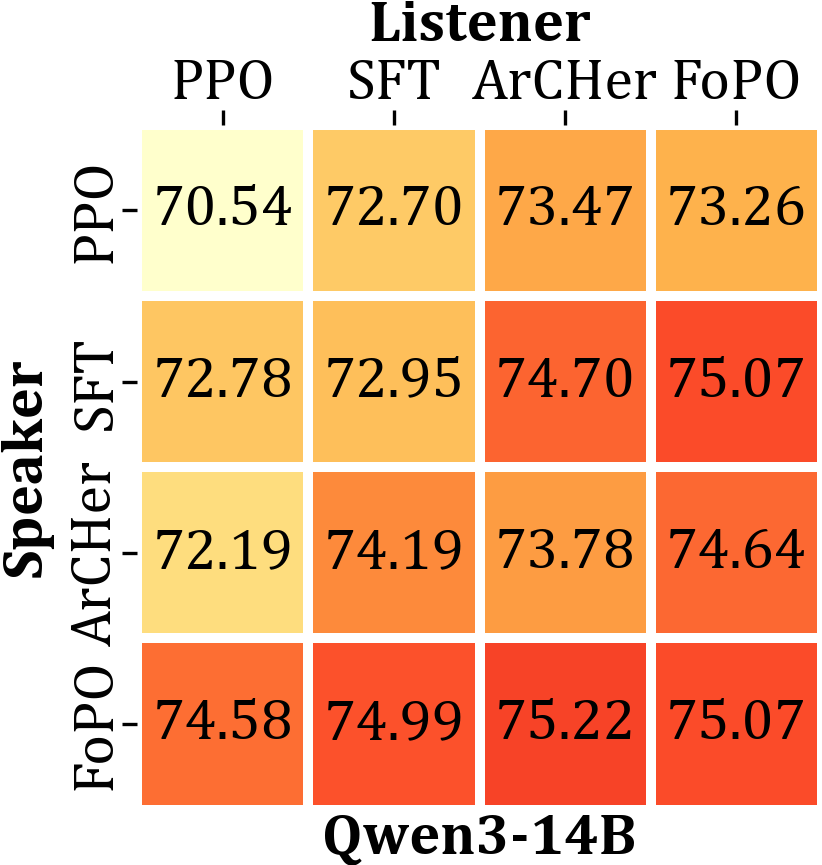}
    \end{minipage}
    \caption{Method performance on \textit{Cooperative RSA}.}
    \label{fig:rsa_reward}
\end{figure}

\begin{figure}[tb]
    \centering
    \begin{minipage}[c]{0.45\textwidth}
        \centering
        \includegraphics[width=\textwidth]{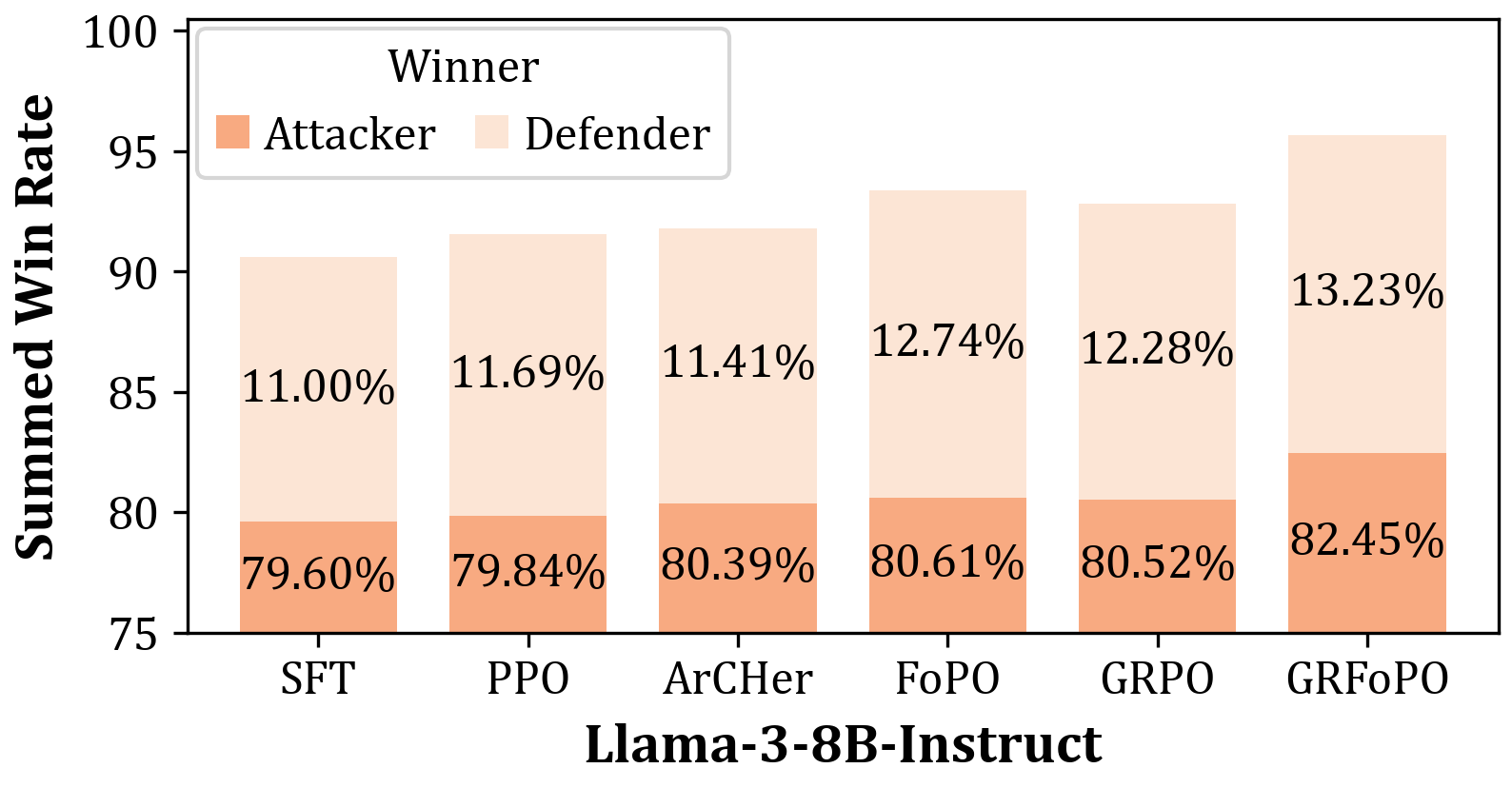}
    \end{minipage}
    \vspace{0.02\textwidth}
    \begin{minipage}[c]{0.45\textwidth}
        \centering
        \includegraphics[width=\textwidth]{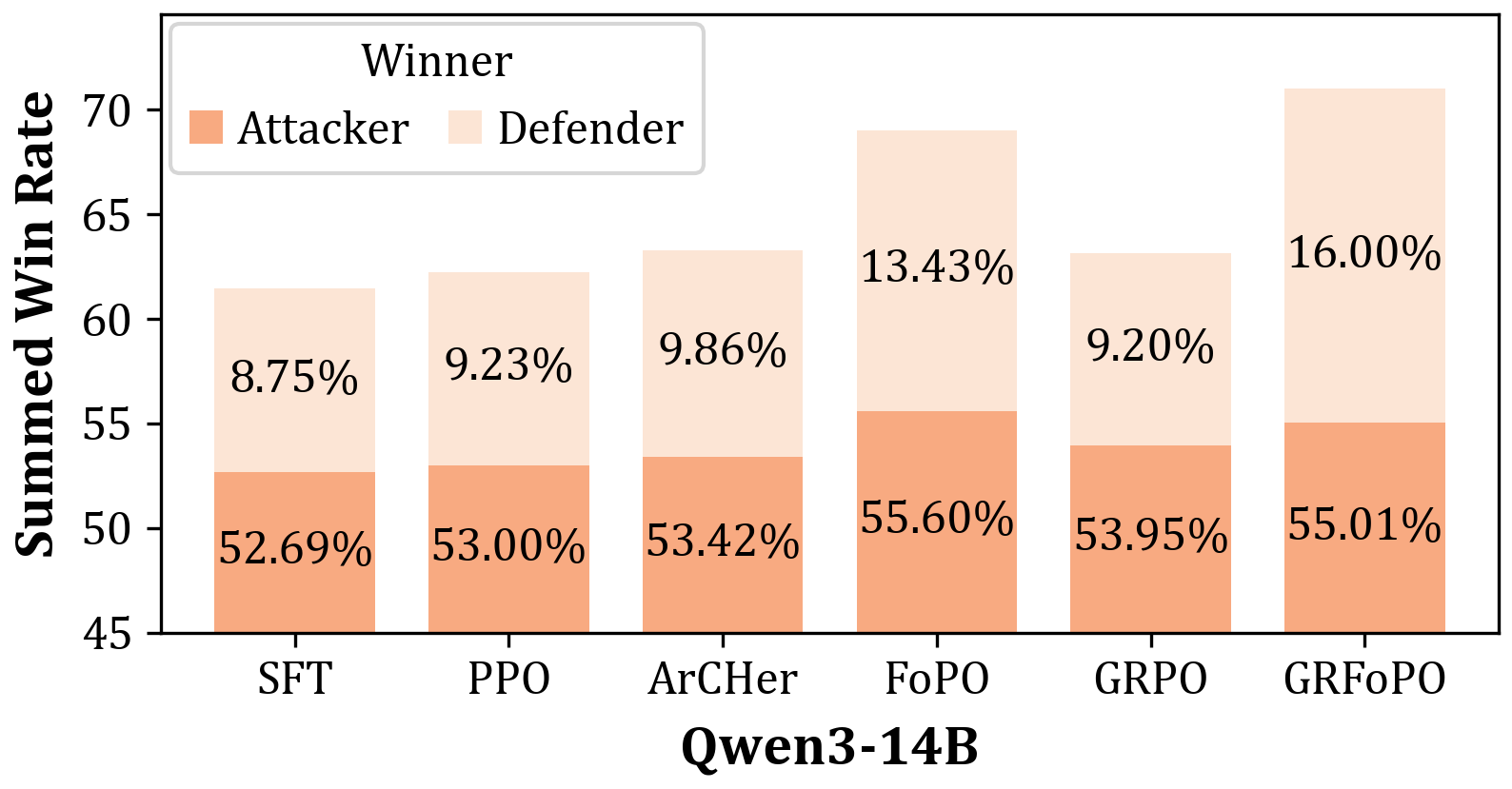}
    \end{minipage}
    \caption{Method performance on \textit{Competitive Taboo}.}
    \label{fig:taboo_winrate}
\end{figure}

\paragraph{In-Domain Evaluation.}
\begin{table*}[tb]
    \centering
    \resizebox{\textwidth}{!}{%
    \begin{tabular}{ccccccccccc}
    \toprule
    \makecell[c]{\textbf{Training} \textbf{Set}} & \textbf{Backbone} & \textbf{Method} & \textbf{Guessing} & \textbf{Bar} & \textbf{Dollar} & \textbf{Diner} & \textbf{Auction} & \textbf{Battle} & \textbf{Pirate} & \textbf{Avg.}\\
    \midrule
    \multirow{9}{*}{RSA} & \multirow{4}{*}{Llama-3-8B-Instruct} & \textit{ICT} & 86.64 & 68.17 & 51.38 & 72.20 & 10.09 & 19.08 & 67.08 & 53.52 \\
    & & \textit{PPO} & 83.16 & 67.67 & 55.23 & \textbf{89.70} & 10.41 & 21.57 & 65.35 & 56.16 \\
    & & \textit{ArCHer} & 87.88 & 67.17 & 58.34 & 88.10 & 10.43 & 19.35 & 66.82 & 56.87 \\
    \cmidrule{3-11}
    & & \textit{FoPO} & \textbf{89.59} & \textbf{67.83} & \textbf{60.36} & 89.10 & \textbf{11.30} & \textbf{29.42} & \textbf{67.40} & \textbf{59.29} \\
    \cmidrule{2-11}
    & \multirow{4}{*}{Qwen3-14B} & \textit{ICT} & 94.24 & 36.33 & 84.24 & 13.70 & 10.61 & 74.66 & 81.85 & 56.52 \\
    & & \textit{PPO} & 94.27 & 49.67 & 81.12 & \textbf{27.00} & 10.21 & 78.75 & 84.12 & 60.73 \\
    & & \textit{ArCHer} & 94.25 & 47.67 & 76.91 & 16.60 & \textbf{10.96} & \textbf{88.04} & 86.07 & 60.07 \\
    \cmidrule{3-11}
    & & \textit{FoPO}  & \textbf{94.87} & \textbf{50.00} & \textbf{88.97} & 20.30 & 10.94 & 82.82 & \textbf{86.15} & \textbf{62.01} \\
    \midrule
    \multirow{13}{*}{Taboo} & \multirow{6}{*}{Llama-3-8B-Instruct} & \textit{ICT} & 83.89 & 66.67 & 59.32 & 95.71 & 16.42 & 16.32 & 44.40 & 54.68 \\
    & & \textit{PPO} & 83.25 & 67.17 & 69.75 & 95.30 & 17.31 & 17.67 & 46.25 & 56.67 \\
    & & \textit{GRPO} & 82.99 & 69.00 & 68.82 & \textbf{96.10} & 20.31 & 18.78 & 47.40 & 57.63 \\
    & & \textit{ArCHer} & 84.48 & 67.33 & 65.81 & 94.70 & 16.70 & 18.05 & 48.62 & 56.53 \\
    \cmidrule{3-11}
    & & \textit{FoPO} & 84.30 & 69.33 & \textbf{71.65} & 95.20 & \textbf{20.84} & 24.60 & 47.96 & 59.13 \\
    & & \textit{GR.FoPO} & \textbf{84.57} & \textbf{71.67} & 66.46 & 94.70 & 18.51 & \textbf{29.09} & \textbf{49.57} & \textbf{59.22} \\
    \cmidrule{2-11}
    & \multirow{6}{*}{Qwen3-14B} & \textit{ICT} &  91.98 & 30.50 & 78.68 & 18.60 & 11.41 & 76.48 & 81.18 & 55.55 \\
    & & \textit{PPO} & 93.83 & \textbf{38.50} & 79.10 & 22.50 & 11.38 & 91.17 & 82.87 & 59.91 \\
     & & \textit{GRPO} & \textbf{93.85} & 34.50 & 80.33 & 22.80 & 11.31 & 89.36 & 82.41 & 59.22 \\
    & & \textit{ArCHer} & 93.71 & 37.00 & 84.42 & 26.70 & 11.06 & 89.00 & 80.29 & 60.31 \\
    \cmidrule{3-11}
    & & \textit{FoPO} & \textbf{93.85} & 35.33 & \textbf{92.30} & \textbf{26.90} & 11.36 & 86.76 & \textbf{85.10} & \textbf{61.66} \\
    & & \textit{GR.FoPO} & 93.62 & 35.33 & 87.25 & 24.70 & \textbf{11.82} & \textbf{93.50} & 84.25 & 61.50 \\
    \midrule
    \multirow{9}{*}{Taboo + RSA} & \multirow{4}{*}{Llama-3-8B-Instruct} & \textit{ICT} & 77.13 & 71.13 & 62.81 & 91.80 & 11.96 & 22.94 & 46.16 & 54.85 \\
    & & \textit{PPO} & 78.29 & 72.00 & 60.99 & 97.80 & 12.51 & 25.80 & 49.58 & 56.71 \\
    & & \textit{ArCHer} & 78.78 & \textbf{73.83} & 57.17 & 93.40 & 10.35 & 21.49 & 46.19 & 54.46 \\
    \cmidrule{3-11}
    & & \textit{FoPO}  & \textbf{80.47} & 72.83 & \textbf{64.61} & \textbf{98.40} & \textbf{13.27} & \textbf{32.94} & \textbf{58.05} & \textbf{60.08} \\
    \cmidrule{2-11}
    & \multirow{4}{*}{Qwen3-14B} & \textit{ICT} & 92.39 & 40.33 & 84.24 & 32.30 & 11.22 & 79.54 & 83.21 & 60.46 \\
    & & \textit{PPO} & 93.88 & 43.83 & 85.79 & 32.40 & 11.73 & 84.00 & 83.07 & 62.10 \\
    & & \textit{ArCHer} & 93.83 & 42.00 & 80.15 & 30.10 & 11.16 & 80.67 & \textbf{85.77} & 60.53 \\
    \cmidrule{3-11}
    & & \textit{FoPO} & \textbf{94.12} & \textbf{52.33} & \textbf{87.85} & \textbf{32.70} & \textbf{11.76} & \textbf{87.29} & 84.04 & \textbf{64.30} \\
    \bottomrule
    \end{tabular}}
    \caption{Cross-method and cross-dataset performance on $\gamma$-Bench.}
    \label{tab:gama_method}
\end{table*}
We evaluate LLM performance by having pairs of models play the game of Cooperative RSA and Competitive Taboo using an evaluation set of 1K instances. For RSA, we report the average conversation reward computed by \Cref{eqn:rsa_reward} and scaled by a factor of 100, averaging results across all model pairings. \Cref{fig:rsa_reward} shows the results. \textbf{(1)} 
Results involving GRPO and GR.FoPO are not reported, as these methods often experience a collapse in token probabilities when trained on RSA, preventing meaningful evaluation; GRPO has also been reported to be less stable than PPO in prior works~\cite{xue2025simpletir,jin2025search}. However, this phenomenon does not occur in Taboo. We hypothesize this stems from differences in return semantics: Taboo returns reflect binary task completion, whereas RSA returns measure continuous cooperation quality. This causes \textit{GRPO's advantages to penalize successful trajectories that achieve sub-optimal rationality erroneously}, as they receive below-mean returns and negative advantages. This issue can be exacerbated by the vast dialogue trajectory space, where limited sampling leads to increased unreliability. \textbf{(2)} Including a model trained with FoPO increases the conversation reward, indicating that FoPO enhances cooperative strategic reasoning in LLMs. Notably, the improvement is slightly more pronounced when the FoPO model assumes the role of the listener. It is because listener-side rational inference plays a decisive role in disambiguating utterances and recovering the speaker's intent, which is also observed in prior work \citep{yuan2018understanding}.

In Taboo, we report each model's win rate as both attacker and defender. From \Cref{fig:taboo_winrate}: \textbf{(1)} Our FoPO and GR.FoPO substantially outperform other methods, demonstrating their effectiveness in enhancing competitive strategic reasoning. \textbf{(2)} The foresight-based correction can be seamlessly integrated into PPO-style methods without compromising their original benefits. This is evident from the fact that GRPO outperforms PPO, and the advantage is maintained when the correction is applied: GR.FoPO continues to outperform FoPO.

\paragraph{Out-of-Domain Evaluation.} 
We adopt the widely used \textbf{$\gamma$-Bench}~\citep{huang2025competing} for out-of-domain evaluation. It is a prompt-based, data-free benchmarking framework designed to assess LLM performance in multi-agent environments through classical game-theoretic scenarios that emphasize strategic interactions and decision-making. We select seven tasks that specifically highlight settings in which agents aim to maximize their individual utility. Each model is evaluated over five runs per task, and we report the average score, following~\cite{huang2025competing}. \Cref{tab:gama_data} presents results for models trained via SFT on different datasets. Higher scores indicate better performance, and \textbf{bold} values highlight the greatest improvements over the corresponding backbone model.
In most tasks, the highest score is achieved by a model trained on our dataset. Moreover, Cooperative RSA demonstrates greater effectiveness than Competitive Taboo, likely due to its stronger emphasis on modeling the counterpart’s reasoning. Notably, models trained on both datasets achieve the best performance across tasks.
Overall, the models trained on our curated datasets consistently achieve better performance, validating that \textbf{our dataset curation contributes to LLMs' strategic reasoning}.

\Cref{tab:gama_method} shows comparison results across algorithms. We observe that ICT's performance drops consistently, indicating that the improvements observed in baseline algorithms are primarily due to reward-based learning rather than patterns in the training data. All RL algorithms improve performance on Competitive Taboo. However, PPO (Llama-3-8B-Instruct) underperforms when trained on Cooperative RSA or both, and ArCHer (Llama-3-8B-Instruct and Qwen3-14B) underperforms when trained on both. This suggests that these methods struggle to effectively leverage the Cooperative RSA reward, which signals the success of cooperation. In contrast, the foresight optimization employs a more effective strategy that accounts for both cooperative and competitive rewards, consistently outperforming other methods. These results demonstrate that \textbf{models trained with foresight optimization consistently exhibit stronger strategic reasoning capabilities, regardless of the training tasks or backbone models used}.

\section{Conclusion}
This work aims to enhance strategic reasoning in LLMs for multi-agent interactions. We identify the foresight nature of strategic reasoning and propose FoPO, a novel algorithm that optimizes for both self-oriented outcomes and anticipated counterpart actions. 
Considering the limitations of existing datasets in strategic reasoning, we curate two new datasets: Cooperative RSA and Competitive Taboo. We finetune LLMs using FoPO in a self-play framework with our datasets. Experimental results prove that our datasets and approach effectively enhance the strategic reasoning capabilities of LLMs.
This work paves the way for more sophisticated, forward-thinking AI systems capable of high-stakes collaboration and competition in real-world scenarios.

\section{Acknowledgment}
We thank the anonymous area chair and anonymous reviewers for their insightful comments and valuable feedback during the review process. This study is funded by the Research Grants Council (project code: T43-518/24-N and PolyU/15213323) under the University Grants Committee, Hong Kong Special Administrative Region Government.

\section*{Limitations}
Our work makes deliberate design choices to enable systematic investigation of the enhancement of strategic reasoning. We focus on linguistic strategic reasoning through dialogue-based tasks, demonstrating that LLMs can develop sophisticated strategic capabilities without relying on external symbolic modules or game-theoretic solvers. Our datasets target fundamental interaction motives, cooperation, and competition, which provide clear training signals while enabling generalization to mixed-motive scenarios, as demonstrated in our out-of-domain evaluation.

Several promising directions for future work emerge. First, extending our framework to incorporate explicit world state representations could enable strategic reasoning in complex multi-agent environments beyond dialogue. Second, while our datasets cover foundational interaction patterns, exploring additional scenarios would further validate the generality of our approach, such as negotiation (balancing cooperation and competition) \cite{hua-etal-2024-assistive}, and multi-party interactions \cite{ki-etal-2025-multiple} (involving mixed motives among multiple agents). Finally, investigating the interplay between strategic reasoning and other cognitive capabilities (e.g., long-term planning \cite{wang2025sparlreinforcingllmagents,song2023llmplanner}, theory of mind \cite{xiao-etal-2025-towards}, mental support \cite{wang-etal-2024-muffin}) presents an exciting research direction.

\bibliography{custom}

\clearpage
\appendix
\startcontents[appendices]
\printcontents[appendices]{}{1}{\section*{Appendix Contents}}
\section{FoPO Derivation}\label{appn:sec:fopo}
\begin{table*}[tb]
\centering
\begin{tabular}{
>{\centering\arraybackslash}m{0.17\linewidth} |
>{\centering\arraybackslash}m{0.38\linewidth} |
m{0.35\linewidth}}
\hline
\textbf{Term}  & \textbf{Representation} & \multicolumn{1}{c}{\textbf{Interpretation}} \\
\hline
$\nabla_{\theta} \mathcal{O}^1(\theta, \theta)$ 
& $\nabla_\theta r_t^{1}(\theta)  \hat{A}^{1,\operatorname{clip}}_{t}$ 
& The standard policy gradient of agent 1, representing local improvement of its own return. \\
\hline
$\nabla_{\theta_2} \mathcal{O}^1(\theta, \theta_2)$ 
& \[
\begin{aligned}
& \nabla_{\theta_2} \left( r_t^{1}(\theta)  r_{t+1}^{2}(\theta)  \hat{A}^{1,\operatorname{clip}}_{t} \right) \\
& = r_t^{1}(\theta)  \nabla_{\theta_2} r_{t+1}^{2}(\theta)  \hat{A}^{1,\operatorname{clip}}_{t} \\
& \xrightarrow{\theta_2 = \theta} r_t^{1}(\theta)  \nabla_{\theta} r_{t+1}^{2}(\theta)  \hat{A}^{1,\operatorname{clip}}_{t}
\end{aligned}
\]
& Measures how agent 1's gain changes in response to updates in agent 2's policy parameters. \\
\hline
$\nabla_{\theta} \nabla_{\theta_2} \mathcal{O}^2(\theta, \theta_2)$ 
& \[
\begin{aligned} 
& \nabla_{\theta} \nabla_{\theta_2} \left[ r_t^{1}(\theta)  r_{t+1}^{2}(\theta)  \hat{A}^{(2,\operatorname{clip}}_{t+1} \right] \\
& = \nabla_{\theta} r_t^{1}(\theta)  \nabla_{\theta_2} r_{t+1}^{2}(\theta)  \hat{A}^{(2,\operatorname{clip}}_{t+1} \\
& \xrightarrow{\theta_2 = \theta} \nabla_{\theta} r_t^{1}(\theta)  \nabla_{\theta} r_{t+1}^{2}(\theta)  \hat{A}^{(2,\operatorname{clip}}_{t+1}
\end{aligned}
\]
& Reflects how agent 2's learning dynamics are influenced by agent 1's policy. \\
\hline
\end{tabular}
\caption{Interpretation of each term in the FoPO update.}
\label{tab:fopo_term_explanation}
\end{table*}
The gain of agent 1 can be represented by the surrogate objective $\mathcal{O}^1(\text{Agent 1}, \text{Agent 2})$, which depends on both agents' strategies. A core idea in FoPO is that agent 1 should take an action that not only maximizes its final reward based on the current generated action, but also anticipates and responds to how agent 2 might change its behavior after its own optimization. Following opponent learning, we aim to maximize the ``foresight gain'':
\begin{align}
\mathcal{O}^1\big(\pi(\text{prompt}^{1}, \theta),\; \pi(\text{prompt}^{2}, \theta + \Delta\theta)\big),
\end{align}
where $\Delta\theta$ is the direction in which agent 2 updates its policy during optimization.

Assuming $\Delta\theta$ is small, we can apply a first-order Taylor expansion with respect to the second argument:

\begin{equation}
\begin{split}
\mathcal{O}^1(\theta,\; \theta + \Delta\theta) 
&\approx \mathcal{O}^1(\theta, \theta) \\
+\; &(\Delta\theta)^\top \nabla_{\theta_2} \mathcal{O}^1(\theta, \theta_2) \Big|_{\theta_2 = \theta},
\end{split}
\end{equation}

where the notation $\mathcal{O}^1(\theta_1, \theta_2)$ represents agent 1's value when the two agents use policy parameters $\theta_1$ and $\theta_2$, respectively. The agent 2's update, given the foresight weight $\eta$, is:

\begin{align}
\Delta\theta = \eta \nabla_{\theta_2} \mathcal{O}^2(\theta, \theta_2) \Big|_{\theta_2 = \theta},    
\end{align}

Substituting this into the Taylor expansion and differentiating $\mathcal{O}^1(\theta, \theta + \Delta\theta)$ with respect to $\theta$, we obtain the FoPO update rule:
\begin{equation}
\begin{split}
\label{eqn:derivation_1}
&\theta_{t+1} 
\leftarrow \theta_t 
+ \alpha  \nabla_{\theta} \mathcal{O}^1(\theta, \theta) \\
& + \alpha\eta  \left( \nabla_{\theta_2} \mathcal{O}^1(\theta, \theta_2) \right)^\top 
\nabla_{\theta} \nabla_{\theta_2} \mathcal{O}^2(\theta, \theta_2) \Big|_{\theta_2 = \theta}.
\end{split}
\end{equation}

Unlike prior opponent modeling approaches~\citep{foerster2018learning, prajapat2021competitive}, we deliberately truncate the dependency of $\nabla_{\theta_2}\mathcal{O}^1(\theta,\theta_2)$ on $\theta_1$. This design choice allows us to focus on the forward influence: how self's current behavior shapes the counterpart's subsequent policy update, and how this adapted policy then affects self's future returns. Instead of differentiating through the opponent's learning rule, we compute this influence via the forward chain rule, avoiding the need to backpropagate through the counterpart's gradient computation and thus eliminating the requirement for mixed Hessian terms $\frac{\partial^2 \mathcal{O}^1}{\partial \theta_1 \partial \theta_2}$. This approach sidesteps the prohibitive computational costs and complexity that characterize previous opponent modeling methods, making it well-suited for enhancing the strategic reasoning of LLM-based agents.

\Cref{tab:fopo_term_explanation} provides the interpretation of each term in \Cref{eqn:derivation_1}. Finally, the FoPO updates the parameters by

\begin{equation}\label{eqn:derivation_2}
\begin{split}
    \theta_{t+1} 
\leftarrow & \theta_t 
+ \alpha  \nabla_\theta r_t^{1}(\theta)  \hat{A}^{1,\operatorname{clip}}_{t} \\
+& \alpha\eta  \underbrace{
\left[
    r_t^{1}(\theta)  \hat{A}^{1,\operatorname{clip}}_{t}  \nabla_{\theta} r_{t+1}^{2}(\theta) 
\right]^\top
}_{\text{Sensitivity of } \mathcal{O}^1 \text{ to } 2} \\
&\quad \underbrace{
\nabla_{\theta} r_t^{1}(\theta)  \nabla_{\theta} r_{t+1}^{2}(\theta)  \hat{A}^{2,\operatorname{clip}}_{t+1}
}_{\text{Effect of 1 on $\mathcal{O}^2$}}.
\end{split}
\end{equation}

When the KL divergence term is included, the formulation becomes equivalent to \Cref{eqn:fopo}. Although FoPO is applicable to agents with differing parameters, our approach focuses on enhancing LLMs' strategic reasoning via self-play.

\section{Cooperative RSA}\label{appn:sec:rsa}
\subsection{Game Rules}
The Cooperative RSA game is a multi-turn interaction between a speaker and a listener. Both agents share an object list $O=\{o_1, \dots, o_N\}$, where each object has $M$ binary-valued features $F=\{f_1, f_2, \dots,f_M \}$. The speaker refers to a target object $\hat{o}$ by revealing one feature per turn, while the listener responds with a subset of objects consistent with the received feature. The game succeeds if the listener isolates $\hat{o}$ as a singleton set and fails if the target is ever excluded. This setup encourages pragmatic reasoning: the speaker must select informative features strategically, and the listener incrementally refines its hypotheses. More efficient interactions, requiring fewer turns, reflect stronger alignment and reasoning capabilities.

\subsection{Bayesian Inference}\label{appn:sec:rsa_bayesian}

The behaviors of rational speakers and listeners are modeled via a Bayesian process. We divide the interaction at the $t$-th and $(t+1)$-th turns into the speaker and listener sides.

\paragraph{Speaker}
At turn $t$, the speaker evaluates each feature $f_m(\hat{o})$ of the target object $\hat{o}$ given candidate objects $O^{(t)}$.
The literal speaker assumes a uniform prior over objects containing the specific feature $f_m(\hat{o})$:

\begin{equation*}
\begin{aligned}
P_{L_1}&(o_n \mid \hat{f}^{(t)}, O) = 
\begin{cases}
{|\hat{f}^{(t)}|^{-1}}, & o_n \in C,\\
0, & \text{otherwise.}
\end{cases}\
\end{aligned}
\end{equation*}
The rational speaker applies Bayesian inference to calculate the posterior for feature $f_m(\hat{o})$
\begin{equation*}
\begin{split}
P_{L_0}(o_n \mid &f_m(\hat{o}), O^{(t)}) = \\
&\frac{P(f_m(\hat{o}) \mid o_n, O^{(t)}) \, P(o_n)}
     {\sum_{o \in O^{(t)}} P(f_m(\hat{o}) \mid o, O^{(t)}) \, P(o)},
\end{split}
\end{equation*}
with likelihood:
\begin{equation*}
    P(f_m(\hat{o}) \mid \hat{o}, O^{(t)}) = \frac{|f_m(\hat{o})|^{-1}}{\sum_{f \in F} |f|^{-1}},
\end{equation*}
where $|f|$ is the number of objects in $O^{(t)}$ possessing feature $f$.

For each feature, the target rank is
\begin{equation*}
\begin{aligned}
\text{rank}_{f_m(\hat{o})}(\hat{o}) =
\left| \left\{ o_n \right. \right. & \in O^{(t)} : \\
P_{L_0}(o_n \mid f_m(\hat{o}), O^{(t)})
& \left. \left. \ge P_{L_0}(\hat{o} \mid f_m(\hat{o}), O^{(t)}) \right\} \right|
\end{aligned}
\end{equation*}
The speaker selects the feature with the highest discriminability:
\begin{equation*}
    \hat{f}^{(t)} = \arg\min_{m=1,\dots,M} \text{rank}_{f_m(\hat{o})}(\hat{o}).
\end{equation*}

\paragraph{Listener}
At turn $(t+1)$, the listener observes $\hat{f}^{(t)}$ and updates its posterior:
\begin{equation*}
\begin{split}
P_{L_1}(o_n &\mid \hat{f}^{(t)}, O^{(t)}) 
\\&= \frac{P(\hat{f}^{(t)} \mid o_n, O^{(t)}) \, P(o_n)}
       {\sum_{o \in O^{(t)}} P(\hat{f}^{(t)} \mid o, O^{(t)}) \, P(o)},
\end{split}
\end{equation*}
with the same likelihood as above.

To model pragmatic inference, the listener simulates the speaker's choice:
\begin{enumerate}
\item For each $o_n \in O^{(t)}$ with $\hat{f}^{(t)} \in o_n$, compute features $F(o_n)$.
\item Simulate the speaker selecting the most informative feature:
\begin{equation*}
f^*_{o_n} = \arg\max_{f \in F(o_n)} P_{L_0}(o_n \mid f, O^{(t)}).
\end{equation*}
\item Retain $o_n$ if $f^*_{o_n} = \hat{f}^{(t)}$.
\end{enumerate}

The listener's belief set $\text{BeliefSet}(\hat{f}^{(t)})$ is formulated as
\[
\left\{ o_n \in O^{(t)} \;\middle|\; 
\hat{f}^{(t)} \in o_n \text{ and } f^*_{o_n} = \hat{f}^{(t)} \right\}.
\]
The next candidate set is $O^{(t+2)}$ is
\[
\arg\max_{o_n \in \text{BeliefSet}(\hat{f}^{(t)})} P_{L_1}(o_n \mid \hat{f}^{(t)}, O^{(t)}).
\]
If only one object remains, it is returned as the final selection.

\subsection{Example in Figure 2}
Consider the example in \Cref{fig:dataset}, where the object set is $O$ = $\{$dry-blue-smooth-square, wet-green-rough-square, wet-green-smooth-square, wet-blue-smooth-circle, wet-blue-smooth-square, dry-blue-rough-circle, dry-blue-rough-square, dry-blue-smooth-circle$\}$, with target object $\hat{o}=o_8=$ dry-blue-smooth-circle. Let the features be moisture, color, texture, and shape.

\paragraph{Speaker Calculation}
First, The speaker evaluates each feature $f_m(\hat{o})$ to choose the most informative one:

\textbf{(1) Moisture = dry}: occurs in $\{$dry-blue-smooth-square, dry-blue-rough-circle, dry-blue-rough-square, dry-blue-smooth-circle$\}$.
\[
P(\text{dry} \mid {o_1}, O) = \frac{\frac{1}{4}}{\frac{1}{4} + \frac{1}{6} + \frac{1}{5} + \frac{1}{5}} = \frac{15}{49} \quad
\]
\[
P(\text{dry} \mid {o_6}, O) = \frac{\frac{1}{4}}{\frac{1}{4} + \frac{1}{6} + \frac{1}{3} + \frac{1}{3}} = \frac{3}{13} \quad
\]
\[
P(\text{dry} \mid {o_7}, O) = \frac{\frac{1}{4}}{\frac{1}{4} + \frac{1}{6} + \frac{1}{3} + \frac{1}{5}} = \frac{15}{57} \quad
\]
\[
P(\text{dry} \mid {o_8}, O) = \frac{\frac{1}{4}}{\frac{1}{4} + \frac{1}{6} + \frac{1}{3} + \frac{1}{5}} = \frac{15}{57} \quad
\]
\[
P(\hat{o_1} \mid \text{dry}, O) = \frac{\frac{15}{49}}{\frac{15}{49}+\frac{3}{13} +\frac{15}{57} + \frac{15}{57} } 
\]
\[
P(\hat{o_6} \mid \text{dry}, O) = \frac{\frac{3}{13}}{\frac{15}{49}+\frac{3}{13} +\frac{15}{57} + \frac{15}{57} } 
\]
\[
P(\hat{o_7} \mid \text{dry}, O) = \frac{\frac{15}{57}}{\frac{15}{49}+\frac{3}{13} +\frac{15}{57} + \frac{15}{57} } 
\]
\[
P(\hat{o_8} \mid \text{dry}, O) = \frac{\frac{15}{57}}{\frac{15}{49}+\frac{3}{13} +\frac{15}{57} + \frac{15}{57} } 
\]
The rank of the target object would be the second, sharing the spot with one other object (dry-blue -rough-square).\\
\textbf{(2) Color = blue}: occurs in $\{$dry-blue-smooth -square,  wet-blue-smooth-circle, wet-blue-smooth -square, dry-blue-rough-circle, dry-blue-rough -square, dry-blue-smooth-circle$\}$.
The ranking of the target object is the third, sharing the spot with one other object (wet-blue-smooth-circle).\\
\textbf{(3) Texture = smooth}: occurs in $\{$dry-blue -smooth-square, wet-green-smooth-square, wet -blue-smooth-circle, wet-blue-smooth-square, dry -blue-smooth-circle$\}$.
The ranking of the target object is third, sharing the spot with one other object (wet-blue-smooth-circle).\\
\textbf{(4) Shape = circle}.
occurs in $\{$dry-blue-smooth-square, wet-green-rough-square, wet-green-smooth-square, wet-blue-smooth-square, dry-blue-rough-square$\}$.
The ranking of the target object is one of the first, sharing the spot with one other object (wet-blue-smooth-circle). \\
By comparing the above rankings, a rational speaker would select the most informative feature, which is ``Shape=circle,'' in the first turn.
\paragraph{Listener Calculation}
At the second turn, the listener observes $\hat{f}^{(t)} = $``Shape=circle.'' First, compute the prior (literal posterior) for each object:
\begin{equation*}
\begin{aligned}
P_{L_1}&(o_n \mid \hat{f}^{(t)}, O) = 
\begin{cases}
{|\hat{f}^{(t)}|^{-1}}, & o_n \in C,\\
0, & \text{otherwise.}
\end{cases}\\
C&=\{\\
&\text{dry-blue-smooth-circle},\\ &\text{wet-blue-smooth-circle},\\ &\text{dry-blue-rough-circle}\\
\}&
\end{aligned}
\end{equation*}
A rational listener simulates the speaker's choice among each candidate in $\{$dry-blue-smooth-circle, wet-blue-smooth-circle, dry-blue-rough-circle$\}$ is as follows:

For the first object \textbf{(1) dry-blue-smooth-circle}:
If the speaker wants to refer to the object dry-blue-smooth-circle, it is most likely that the speaker would say ``circle.'' ``Circle'' matches the speaker's utterance. Therefore, dry-blue-smooth-circle is retained in $\text{BeliefSet}(\hat{\text {dry}}^{(t)})$.
Refer to the previous section for the calculation process.
\textbf{(2) wet-blue-smooth-circle:} Similar to the previous calculation, to refer to this object, the speaker should have said ``circle.'' ``Circle'' matches the speaker's utterance. Therefore, wet-blue-smooth-circle is retained in $\text{BeliefSet}(\hat{\text {dry}}^{(t)})$.
\textbf{(3) dry-blue-rough-circle:} Similar to the previous calculation, to refer to this object, the speaker should have said ``rough.'' ``Rough'' does not match the speaker's utterance. Therefore, dry-blue-rough-circle is not retained in $\text{BeliefSet}(\hat{\text {dry}}^{(t)})$.
Then, the candidate set is
\begin{equation}
\begin{aligned}
    O^{(t+2)} &= \text{BeliefSet}(\hat{\text {circle}}^{(t)}) \\
        &= \{\text{dry-blue-smooth-circle}, \\
        & \text{wet-blue-smooth-circle}\}.
\end{aligned}
\end{equation}

\subsection{Speaker \& Listener Foresight}
\begin{figure}[tb]
\begin{center}
\includegraphics[width=.7\linewidth]{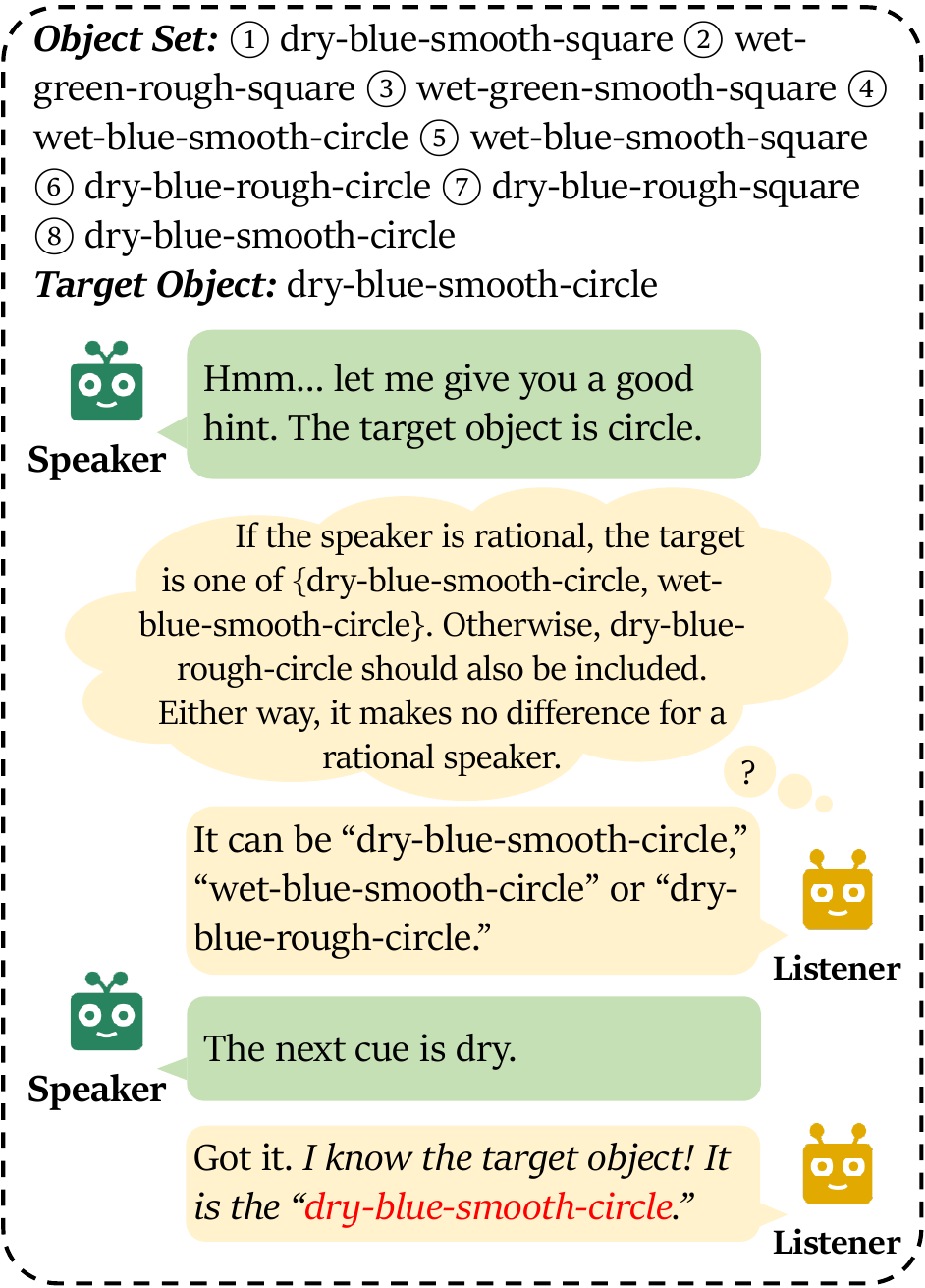}
\end{center}
\caption{PPO optimized the self policy ($\pi_1$) in isolation, while FoPO introduces foresight into the future updates of the counterpart Policy ($\pi_2$).}
\label{appn:fig:listener_foresight}
\end{figure}

In the cooperative RSA game, the speaker's foresight can be naturally modeled via Bayesian inference as formulated in RSA, whereas the listener's foresight is more complex. As described in \Cref{appn:sec:rsa_bayesian}, a rational listener reasons based solely on the speaker's behavior in the previous turn, implicitly assuming the speaker is rational. In practice, however, the listener cannot be certain of the speaker's rationality and must therefore account for multiple possible courses of the speaker's future actions. This process is illustrated in \Cref{appn:fig:listener_foresight}.

\subsection{Data Construction}
The data construction pipeline can be summarized as: (Feature Pair Bank Construction, Objective Matrix and Object Construction) $\rightarrow$ Dialogue Chain Computation $\rightarrow$ LLM-based Dialogue Generation. Each step is illustrated as follows:
\paragraph{Feature Pair Bank Construction}
We first construct a curated set of binary feature pairs, each representing a minimal semantic contrast (e.g., \textit{smooth} vs. \textit{rough}, \textit{graceful} vs. \textit{clunky}). This bank is partitioned into two disjoint subsets to prevent data leakage between training stages.
\begin{itemize}
    \item \textbf{SFT Feature Pair Bank:} A set of 86 pairs used to generate polished dialogue for supervised fine-tuning.
    \item \textbf{RL Feature Pair Bank:} A set of 25 pairs used exclusively to construct ranking-based preference data for reinforcement learning.
\end{itemize}
This separation ensures a clean experimental boundary between learning phases, as the RL component does not optimize on features the model has already seen during supervised training.

\paragraph{Objective Matrix and Object Construction}
Each matrix in our system encodes a semantic mapping between feature dimensions and a set of candidate referents. An entry of 1 indicates that a referent shares the same value as the target referent for a given feature, while 0 denotes a mismatch.
$$
M_{i,j} = \begin{cases}
1, & \text{referent } i \text{ matches target on feature } j \\
0, & \text{otherwise}
\end{cases}
$$
We generate a large pool of such binary matrices with varying shapes, denoted as $m \times n$, where $n$ is the number of candidate referents and $m$ is the number of features. To evaluate the reasoning depth required to resolve each matrix, we simulate golden dialogues using Rational Speaker and Listener models. This allows us to annotate each matrix with the number of rounds required to uniquely identify the target referent through pragmatic inference. These selected matrices, along with features from the feature pair bank, are then used to construct the object list and specify the target object for each dialogue task.

\paragraph{Dialogue Chain Computation}
Using the constructed object list and the target object, we employ the RSA model, illustrated in \Cref{appn:sec:rsa_bayesian}, to compute the optimal dialogue chain. This process involves iterative pragmatic inference, where a rational speaker chooses an utterance that maximally reduces the listener's uncertainty about the target object, and a rational listener updates their belief distribution accordingly. The output is a sequence of features and object sets updates representing the most efficient path to identifying the target.

\paragraph{LLM-based Dialogue Generation}
The final step is to use an LLM to translate the computed dialogue chain into a natural, conversational format. The LLM takes the structured output of the Bayesian computation as input and generates a realistic dialogue that mirrors the pragmatic choices and reasoning depth of the chain, thereby creating a rich dataset for training and evaluation. In this process, we employed four prompts, with one representative example shown below:

\begin{tcolorbox}[
    enhanced,
    colback=white, colframe=black!20, 
    boxrule=0.2mm,
    arc=0.5mm, 
    fontupper=\ttfamily\scriptsize,
    fonttitle=\scriptsize\bfseries, 
    breakable
]
\textbf{\underline{Representative RSA Conversation Generation Prompt:}}\\\\
You're awesome at making dialogue sound natural and conversational! I need your help turning this robotic dialogue into something that feels like real people chatting.
\\\\
Scenario Overview:

- This is a guessing game: the Speaker describes an object, and the Listener tries to guess what it is.

- The target object the Speaker is referring to is: $\{\text{target\_referent}\}$.

- The Listener needs to figure out what object the Speaker means, using this format when they finally guess: ``I know the target object. It is ...''

- Here are all the possible objects being referred to: $\{ \text{referent\_set} \}$.
\\\\
Original dialogue:

$\{\text{dialogue}\}$
\\\\
This dialogue serves as the backbone of your refined version. Your task is to revise it to a real-world conversation, while maintaining the basic contents: the feature or the object(s).
\\\\
Transform the original dialogue to sound friendly, casual, and human, while keeping the structure and meaning the same. Instructions for the generated dialogue:

1. Keep the same number of lines, turns, and speakers as the original.

2. Each casual line must match the original's meaning and content, just in a more natural tone.

3. Make it sound like real people chatting—relaxed, informal, and friendly.

4. Use casual phrases, natural pauses, filler words (like ``um,'' ``you know''), and everyday language.

5. Keep each line around 70 words—brief, but with a conversational feel.
\\\\
Output Format:

Just give me the improved dialogue in this exact format:

Speaker: [Casual version]

Listener: [Casual version]

Speaker: [Casual version]

Listener: [Casual version]

...

\end{tcolorbox}

\newcommand{\minconv}{10}    
\newcommand{\numfeature}{15} 
\begin{figure}[th]
\centering
\begin{tikzpicture}
\begin{axis}[
    width=8cm,
    height=6cm,
    xlabel={conv\_turn $t_{\text{conv}}$},
    ylabel={$R_{\text{RSA}}$},
    domain=\minconv:\numfeature,
    samples=300,
    ymin=0, ymax=1.05,
    grid=major,
    thick,
    legend pos=south west,
    font=\footnotesize, 
    xtick={\minconv, \numfeature},
    xticklabels={$\text{min\_conv}$, $\text{\#(features)}$} 
]

\addplot[
    blue,
    thick
]
{((x - \numfeature)/(\minconv - \numfeature))^1};
\addlegendentry{$\gamma = 1$}

\addplot[
    red,
    thick,
    dashed
]
{((x - \numfeature)/(\minconv - \numfeature))^2};
\addlegendentry{$\gamma = 2$}

\addplot[
    green!60!black,
    thick,
    dotted
]
{((x - \numfeature)/(\minconv - \numfeature))^(0.5)};
\addlegendentry{$\gamma = 0.5$}

\end{axis}
\end{tikzpicture}
\caption{Higher $\gamma$ leads to stronger penalties for exceeding optimal turns.}
\label{appn:fig:gama}
\end{figure}
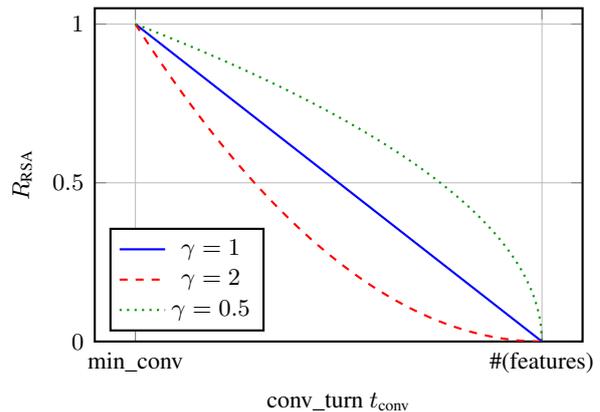
\subsection{Game Reward}
The reward in Cooperative RSA is strongly affected by the parameter $\gamma$, as shown in \Cref{appn:fig:gama}. Values of $\gamma>1$ strongly encourage agents to complete the game in fewer turns, whereas $\gamma<1$ offers a more moderate incentive.

\section{Experiment}\label{appn:sec:training}
\begin{table*}[tb]
    \centering
    \resizebox{\textwidth}{!}{%
    \begin{tabular}{ccccccccccc}
    \toprule
    \textbf{Backbone} & \makecell[c]{\textbf{Training} \textbf{Set}} & \textbf{Method} & \textbf{Guessing} & \textbf{Bar} & \textbf{Dollar} & \textbf{Diner} & \textbf{Auction} & \textbf{Battle} & \textbf{Pirate} & \textbf{Avg.}\\
    \midrule
    \multirow{4}{*}{Llama-3-8B-Instruct} & \multirow{2}{*}{20 Questions} & \textit{SFT} & 76.25 & 63.67 & \textbf{64.99} & 85.50 & 20.49 & 19.59 & 55.81 & 55.19 \\
    & & \textit{PPO} & 78.67 & 64.35 & 64.89 & 87.34 & 27.89 & \textbf{13.18} & 57.30 & 56.23 \\
    & & \textit{FoPO} & \textbf{79.59} & \textbf{65.00} & 64.81 & \textbf{91.00} & \textbf{28.24} & 11.62 & \textbf{58.26} & \textbf{56.93} \\
    \cmidrule{2-11}
    & \multirow{2}{*}{Guess My City} & \textit{SFT} & 90.28 & 53.17 & 53.28 & 91.30 & 26.17 & 15.54 & 43.88 & 53.37 \\
    & & \textit{PPO} & \textbf{90.89} & 55.83 & 53.39 & 91.50 & \textbf{26.18} & 14.19 & 42.58 & 53.51 \\
    & & \textit{FoPO} & 89.79 & \textbf{55.92} & \textbf{55.37} & \textbf{91.71} & 26.13 & \textbf{17.65} & \textbf{45.67} & \textbf{54.61} \\
    \midrule
    \multirow{4}{*}{Qwen3-14B} & \multirow{2}{*}{20 Questions} & \textit{SFT} & 93.66 & \textbf{36.83} & \textbf{77.79} & 6.40 & \textbf{10.93} & 78.14 & 81.88 & 55.09\\
    & & \textit{PPO} &  93.66 & 35.41 & 74.13 & 8.00 & 10.10 & 83.11 & 85.33 & 55.68 \\
    & & \textit{FoPO} & \textbf{94.59} & 33.50 & 77.86 & \textbf{9.40} & 10.87 & \textbf{86.90} & \textbf{92.17} & \textbf{57.90} \\
    \cmidrule{2-11}
    & \multirow{2}{*}{Guess My City} & \textit{SFT} & 94.60 & 36.67 & 71.55 & 9.10 & 11.04 & 80.00 & 83.70 & 55.24 \\
    & & \textit{PPO} & \textbf{94.70} & 38.48 & 77.18 & 8.70 & 10.69 & \textbf{82.32} & 83.17 & 56.46 \\
    & & \textit{FoPO} & 94.69 & \textbf{42.00} & \textbf{94.52} & \textbf{9.20} & \textbf{11.24} & 76.58 & \textbf{84.07} & \textbf{58.90} \\
    \bottomrule
    \end{tabular}}
    \caption{PPO and FoPO Performance on $\gamma$-Bench with deductive reasoning training}
    \label{appn:tab:gama_method}
\end{table*}

\begin{table*}[tb]
    \centering
    \resizebox{0.75\textwidth}{!}{%
    \begin{tabular}{ccccccccc}
    \toprule
    \textbf{Backbone} & \textbf{Guessing} & \textbf{Bar} & \textbf{Dollar} & \textbf{Diner} & \textbf{Auction} & \textbf{Battle} & \textbf{Pirate} & \textbf{Avg.}\\
    \midrule
    GPT-4.1 & 95.15 & 35.83 & 93.12 & 24.80 & 13.24 & 35.00 & 94.92 & 56.01 \\
    DeepSeek-V3.2 & 94.83 & 26.67 & 96.65 & 5.72 & 13.26 & 64.68 & 99.90 & 57.39 \\
    \bottomrule
    \end{tabular}}
    \caption{GPT-4.1 and DeepSeek-V3.2 Performance on $\gamma$-Bench}
    \label{appn:tab:close_gama}
\end{table*}

\subsection{Comparison Models}
To prove the effectiveness of FoPO, we consider the following comparison methods:
(1) \textbf{ICT}~\citep{an2023context}: In context learning augments each training instance with \textit{k} in-context demonstrations and trains with next-token cross-entropy, distilling demonstration patterns into the model parameters. This baseline is included to assess whether the observed improvements stem from learning from rewards or from conversation patterns in the training data. We include this baseline to prove that SFT models are fully trained.
(2) \textbf{PPO}~\citep{Schulman2017PPO}: We include PPO since our FoPO method builds upon it, allowing us to isolate the effect of the foresight-oriented correction in FoPO.
(3) \textbf{ArCHer}~\citep{zhou2024archer}: This is a hierarchical RL algorithm, where a high-level RL algorithm is used to train a value function that aggregates rewards over entire utterances and a low-level RL algorithm then leverages this high-level value function to train a token-by-token policy. Due to the high-level RL, the model can plan across utterances and guide the low-level policy with broader conversational objectives. We include it to compare its explicit long-term planning capability with the explicit counterpart foresight offered by FoPO.
(4) \textbf{GRPO}~\citep{deepseek-math}: We further apply the foresight optimization on top of GRPO to demonstrate the generalizability of our proposed method.

\subsection{Training Datasets}
When fine-tuning models on 20 Questions and GuessMyCity, we randomly sampled 2,300 conversations from each dataset to ensure a comparable number of training steps, and additional 240 instances (keywords or cities) for further RL in both tasks for the same purpose.

\subsection{Training Details}\label{sec:training_details}
To improve training efficiency, we applied LoRA~\citep{hu2022lora} during SFT and subsequently merged the LoRA parameters into the backbone model. We set the rank as 8, alpha as 16, and applied the LoRA modules to the query and value projection layers.
We trained the models using the AdamW optimizer. For SFT, we set the learning rate $\alpha$ to $5\times 10^{-5}$, the KL regularization coefficient $\beta$ to $0.01$, and the batch size to $32$. For RL and ICT, the learning rate $\alpha$ was $1\times 10^{-5}$, $\beta$ was $0.1$, $\eta$ was $0.1$, the reward decay factor $\delta$ is $0.8$, and the batch size was $16$. 
Training was performed on $4$ NVIDIA 5880 (48GB) GPUs. We employ DeepSpeed ZeRO Stage 2~\citep{rajbhandari2020zero} to optimize memory usage and accelerate training. For GRPO, we sample four trajectories for each instance. For PPO and FoPO, we follow the GRPO training protocol and do not train a critic model. This design choice enhances training efficiency and facilitates a more equitable comparison across various policy optimization methods. 

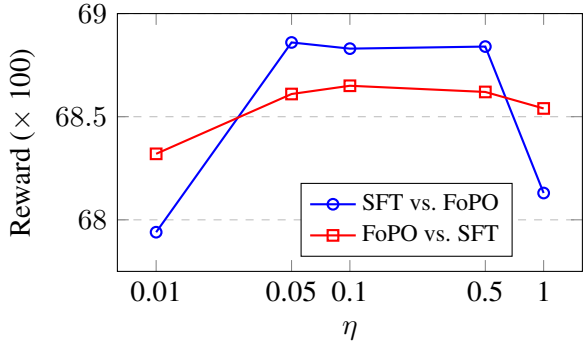
\begin{figure}[t]
    \centering
    \begin{tikzpicture}
        \begin{axis}[
            xlabel={$\eta$},
            ylabel={Reward ($\times$ 100)},
            xmode=log,
            log ticks with fixed point,
            xtick={0.01, 0.05, 0.1, 0.5, 1},
            xticklabels={0.01, 0.05, 0.1, 0.5, 1},
            ymin=67.75, ymax=69,
            ymajorgrids=true,
            grid style=dashed,
            legend style={font=\small, at={(0.85,0.2)}, anchor=east},
            legend style={font=\small},
            width=\linewidth,
            height=5cm,
        ]
        \addplot[
            color=blue, mark=o, thick
        ] coordinates {
            (0.01, 67.94)
            (0.05, 68.86)
            (0.1,  68.83)
            (0.5,  68.84)
            (1,    68.13)
        };
        \addlegendentry{SFT vs.\ FoPO}

        \addplot[
            color=red, mark=square, thick
        ] coordinates {
            (0.01, 68.32)
            (0.05, 68.61)
            (0.1,  68.65)
            (0.5,  68.62)
            (1,    68.54)
        };
        \addlegendentry{FoPO vs.\ SFT}
        \end{axis}
    \end{tikzpicture}
    \caption{Hyperparameter sensitivity of FoPO to the foresight weight $\eta$ on cooperative RSA (reward). Performance is relatively stable across $\eta \in [0.05, 0.5]$.}
    \label{appn:fig:eta_rsa}
\end{figure}
\begin{figure}[t]
\centering
\begin{tikzpicture}
    \begin{axis}[
        name        = topax,
        width       = \linewidth,
        height      = 3.5cm,
        xmode       = log,
        log ticks with fixed point,
        xtick       = {0.01, 0.05, 0.1, 0.5, 1},
        xticklabels = {},
        ymin=67, ymax=73,
        ytick       = {70, 72, 74},
        ymajorgrids = true,
        grid style  = dashed,
        axis x line*= top,
        axis y line*= left,
        legend pos  = north east,
        legend style={font=\small, at={(0.85,0.05)}, anchor=east},
        clip        = false,
    ]
    \addplot[color=red, mark=square, thick] coordinates {
        (0.01, 69.62)(0.05, 69.18)(0.1, 72.15)(0.5, 70.04)(1, 70.14)
    };
    \addlegendentry{FoPO vs.\ SFT}

    \addlegendimage{color=blue, mark=o, thick}
    \addlegendentry{SFT vs.\ FoPO}

    \draw (rel axis cs:1,0) -- (rel axis cs:1,1);

    \fill[white] (rel axis cs:0, 0) +(-4pt,-4pt) rectangle +(4pt,4pt);
    \draw[thin]  (rel axis cs:0, 0) +(-3pt,-4pt) -- +(3pt,2pt);
    \draw[thin]  (rel axis cs:0, 0) +(-3pt,-2pt) -- +(3pt,4pt);
    \end{axis}

    \begin{axis}[
        name        = botax,
        at          = {(topax.south west)},
        anchor      = north west,
        yshift      = 0pt,
        width       = \linewidth,
        height      = 3.5cm,
        xlabel      = {$\eta$},
        xmode       = log,
        log ticks with fixed point,
        xtick       = {0.01, 0.05, 0.1, 0.5, 1},
        xticklabels = {0.01, 0.05, 0.1, 0.5, 1},
        ymin=14, ymax=22,
        ytick       = {14, 16, 18, 20},
        ymajorgrids = true,
        grid style  = dashed,
        axis x line*= bottom,
        axis y line*= left,
        clip        = false,
    ]
    \addplot[color=blue, mark=o, thick] coordinates {
        (0.01, 15.46)(0.05, 18.07)(0.1, 19.68)(0.5, 18.84)(1, 18.07)
    };

    \draw (rel axis cs:1,0) -- (rel axis cs:1,1);

    \fill[white] (rel axis cs:0, 1) +(-4pt,-4pt) rectangle +(4pt,4pt);
    \draw[thin]  (rel axis cs:0, 1) +(-3pt,-4pt) -- +(3pt,2pt);
    \draw[thin]  (rel axis cs:0, 1) +(-3pt,-2pt) -- +(3pt,4pt);
    \end{axis}

    \node[rotate=90, anchor=center]
        at ($(topax.west)!0.5!(botax.west) + (-30pt, 0)$)
        {Win Rate (\%)};

\end{tikzpicture}
\caption{Hyperparameter sensitivity of FoPO to the foresight weight $\eta$ on competitive Taboo (win rate). $\eta = 0.1$ achieves the best performance.}
\label{appn:fig:eta_taboo}
\end{figure}
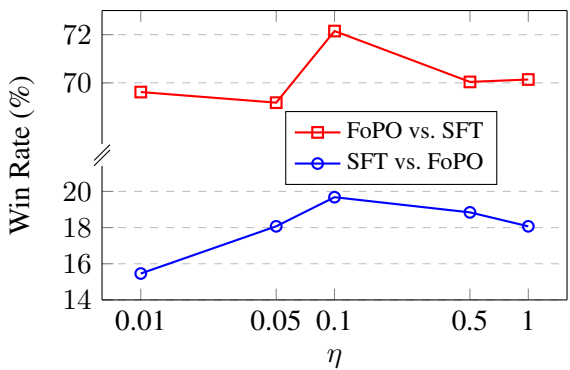

To determine the most suitable value for $\eta$, we evaluated several candidates using a smaller RL training dataset and Meta-Llama-3-8B-Instruct SFT models. Specifically, we trained different FoPO variants with varying $\eta$ values. With the SFT model serving as the counterpart policy, we evaluated each variant via reward on cooperative RSA and win rate on competitive Taboo. As shown in \Cref{appn:fig:eta_rsa} and \Cref{appn:fig:eta_taboo}, $\eta = 0.1$ achieves the best performance across both tasks and was therefore adopted in all experiments.

\begin{table*}[ht]
    \centering
    \small
    \begin{tabular}{>{\RaggedRight\bfseries\arraybackslash}p{0.175\textwidth}|>{\arraybackslash}p{0.77\textwidth}}
    \hline
    \multicolumn{2}{l}{\textbf{Target Object}: loud-weak-late-lean} \\
    \hline
    \multicolumn{2}{p{0.9\textwidth}}{\textbf{Object List}: loud-weak-late-lean, quiet-strong-early-fat, quiet-strong-early-lean, quiet-strong-late-fat, quiet-strong-late-lean, loud-weak-early-fat, loud-weak-early-lean, loud-weak-late-fat}\\
    \hline
    \multicolumn{2}{p{0.9\textwidth}}{\textbf{Chain}: [Speaker: loud, Listener: loud-weak-late-lean, loud-weak-early-fat, loud-weak-early-lean, loud-weak-late-fat, Speaker: late, Listener: loud-weak-late-lean, loud-weak-late-fat, Speaker: lean, Listener: loud-weak-late-lean]} \\
    \hline
    \multicolumn{2}{c}{\textbf{Conversation}} \\
    \hline
    \textbf{Role}    &  \textbf{Turn} \\
    \hline
    Speaker & To initiate the identification process, I would like to highlight the attribute “loud” as a salient feature of the target object. Please consider all candidate objects that possess this particular characteristic as we proceed with the elimination and selection process.\\
    \hline
    Listener & In response to your specification, I have narrowed the potential candidates to the following objects, each of which is characterized by the attribute “loud”: ('loud, weak, late, lean'), ('loud, weak, early, fat'), ('loud, weak, early, lean'), and ('loud, weak, late, fat').\\
    \hline
    Speaker & Building upon the previous description, I would now like to specify that the target object is further distinguished by the attribute “late.” Kindly refine your consideration to those objects which simultaneously exhibit both “loud” and “late” as defining features. \\
    \hline
    Listener & Taking into account your most recent clarification, I have reduced the set of plausible candidates to two objects: ('loud, weak, late, lean') and ('loud, weak, late, fat'), both of which possess the attributes “loud” and “late.”\\
    \hline
    Speaker & To further delineate the target object, I would like to introduce the attribute “lean” as a critical distinguishing characteristic. This should enable you to identify the precise object among the remaining options.\\
    \hline
    Listener & Based on the cumulative information you have provided—specifically, the attributes “loud,” “late,” and “lean”—I am confident in my assessment. I know the target object. It is ('loud, weak, late, lean').\\
    \hline
    \end{tabular}
    \caption{A data instance from the Cooperative RSA dataset.}
    \label{appn:tab:rsa_data}
\end{table*}

\subsection{Code Details}
To implement FoPO, the typical \texttt{backward()} pass is replaced by four separate calls to \texttt{torch.autograd.grad} to compute and assign gradients for each parameter via \texttt{p.grad = ...}. By clearing the computational graph's cache after each call, we ensure that GPU memory usage remains on par with a standard PPO implementation. FoPO's total training time is roughly 3–3.5× that of PPO due to the multiple gradient computations. Compared to PPO, the training time for ArCHer is about eight times greater. Its memory usage also fluctuates, with a peak consumption that is approximately twice PPO's. The datasets themselves will be released upon acceptance of this work.

\subsection{GRPO Fails on Cooperative RSA}
Since we observed that GRPO fails on Cooperative RSA, we investigate whether removing standard deviation normalization, following DR.GRPO \cite{liu2025understanding}, could mitigate the issue. However, DR.GRPO exhibits identical failure modes: models still collapse to deterministically outputting a single token (probability 1.0) at all positions. This indicates the core problem lies in the mean-based advantage computation, which becomes unreliable given the vast dialogue space and limited sampling.

\subsection{Out-of-Domain Evaluation}
We evaluate FoPO on deductive reasoning tasks. Specifically, PPO and FoPO were trained on 20 Questions and Guess My City datasets, and their performance was assessed using $\gamma$-Bench. The results are presented in \Cref{appn:tab:gama_method}. The results indicate that, although FoPO is primarily designed for strategic reasoning, it can also enhance model performance when trained on deductive reasoning tasks. This highlights the ability of foresight optimization to generalize across multiple task types. 

For reference, \Cref{appn:tab:close_gama} reports the performance of GPT-4.1 and DeepSeek-V3.2, the closed-source models used for dataset construction. Their results are comparable to or slightly below those of the SFT models, yet consistently higher than those of the vanilla models (\Cref{tab:gama_data}). This suggests that the structured rules in our data collection pipeline improve dataset quality beyond what purely distilled data would yield.

\section{Case Study}
\subsection{Dataset Cases}
\Cref{appn:tab:rsa_data} presents a sample from the Cooperative RSA dataset, including the conversation chain, with a minimum of three rounds in the conversation. \Cref{appn:tab:taboo_data} presents a sample from the Competitive Taboo dataset, where the target word is ``one'' and the winner is the defender.

\begin{table}[ht]
    \centering
    \small
    \begin{tabular}{>{\RaggedRight\bfseries\arraybackslash}p{0.15\linewidth}|>{\arraybackslash}p{0.725\linewidth}}
    \hline
    \multicolumn{2}{l}{\textbf{Target Word}: One} \\
    \hline
    \multicolumn{2}{c}{\textbf{Conversation}} \\
    \hline
    \textbf{Role}    &  \textbf{Turn} \\
    \hline
    Attacker & If you had to choose a number that symbolizes unity or the beginning of a sequence, which would you pick? \\
    \hline
    Defender & I would probably select something straightforward, perhaps the initial digit used when counting. \\
    \hline
    Attacker & That makes sense! Often, when people start counting out loud, what's the first thing they typically say? \\
    \hline
    Defender & When people start counting out loud, they usually say the very first numeral, often used as the starting point for sequences.\\
    \hline
    Attacker & Consider numbers that are often viewed as singular or representing individuality—what digit comes to mind first?\\
    \hline
    Defender & I know the word! It is ``one!''\\
    \hline
    \end{tabular}
    \caption{A data instance from the Competitive Taboo dataset.}
    \label{appn:tab:taboo_data}
\end{table}

\subsection{FoPO Generation Cases}
The generated conversations from the Cooperative RSA and Competitive Taboo tasks, produced by FoPO Llama-3-8B-Instruct, are shown in \Cref{appn:tab:fopo_rsa} and \Cref{appn:tab:fopo_taboo}. In the Cooperative RSA task, the speaker and listener succeed within the minimum conversation turns. In the Competitive Taboo task, the defender won by successfully guessing the target word. 
\begin{table}[th]
    \centering
    \small
    \begin{tabular}{>{\RaggedRight\bfseries\arraybackslash}p{0.15\linewidth}|>{\arraybackslash}p{0.725\linewidth}}
    \hline
    \multicolumn{2}{l}{\textbf{Target Object}: muted-rough-soggy-no-bumpy} \\
    \hline
    \multicolumn{2}{p{0.9\linewidth}}{\textbf{Object List}: muted-rough-soggy-no-bumpy, echoing-smooth-crisp-yes-flat, echoing-smooth-crisp-no-bumpy, echoing-smooth-soggy-yes-bumpy, muted-rough-crisp-no-bumpy}\\
    \hline
    \multicolumn{2}{c}{\textbf{Conversation}} \\
    \hline
    \textbf{Role} & \textbf{Turn} \\
    \hline
    Speaker & I'm thinking of something that feels rough to the touch.\\
    \hline
    Listener & Okay, so that could be either the one that's muted, rough, soggy, and no, or the one that's muted, rough, crisp, and no, but bumpy. \\
    \hline
    Speaker & And to narrow it down, it's definitely soggy. \\
    \hline
    Listener & I know the target object. It is muted, rough, soggy, and no, bumpy.\\
    \hline
    \end{tabular}
    \caption{FoPO (Llama-3-8B-Instruct) generated Cooperative RSA conversation.}
    \label{appn:tab:fopo_rsa}
\end{table}

\begin{table}[th]
    \centering
    \small
    \begin{tabular}{>{\RaggedRight\bfseries\arraybackslash}p{0.15\linewidth}|>{\arraybackslash}p{0.725\linewidth}}
    \hline
    \multicolumn{2}{l}{\textbf{Target Word}: Appetency} \\
    \hline
    \multicolumn{2}{c}{\textbf{Conversation}} \\
    \hline
    \textbf{Role}    &  \textbf{Turn} \\
    \hline
    Attacker & Let's begin. Imagine you're at a restaurant, and you're trying to decide what to order. What's the first thing you think about? \\
    \hline
    Defender & I think about what type of food I'm in the mood for.\\
    \hline
    Attacker & That's a great point! Now, what if you were at a buffet and you're scanning the options? What's the first thing you consider when choosing what to eat? \\
    \hline
    Defender & I consider what's visually appealing and what looks fresh.\\
    \hline
    Attacker & That's a good strategy. Now, let's think about the first stage of the digestive process. What's the term for the desire or craving for food? \\
    \hline
    Defender & I know the word! It is ``appetency.''\\
    \hline
    \end{tabular}
    \caption{FoPO (Llama-3-8B-Instruct) generated Competitive Taboo conversation, where the FoPO Defender is the winner.}
    \label{appn:tab:fopo_taboo}
\end{table}

\section{LLM Agent Prompts}\label{appn:sec:prompt}
\subsection{Task Prompts}
For each task, we prepare a set of prompts, from which a single prompt is randomly selected during training. Notably, the prompts for Competitive Taboo are also used to generate conversations for the corresponding dataset. We provide several representative prompts here, while the full set is included in the code.

For the Cooperative RSA task, the following prompt was among those used in the implementation:
\begin{tcolorbox}[
    enhanced,
    colback=white, colframe=black!20, 
    boxrule=0.2mm,
    arc=0.5mm, 
    fontupper=\ttfamily\scriptsize,
    fonttitle=\scriptsize\bfseries,
    breakable
]
\textbf{\underline{Representative Cooperative RSA Task Prompt:}}\\\\
Embark on the collaborative challenge of the Rational Speech Act Game, where players assume the roles of either speaker or listener.
\\\\
The speaker enters the game with a covertly assigned target object, while the listener starts without knowledge of this object. The speaker's goal is to effectively guide the listener toward identifying the target object, thereby securing victory. However, there's a rule: the speaker may only provide one feature per turn.
\\\\
Simultaneously, the listener's task is to deduce the target object and present possible target referent objects at each turn. The listener benefits from the ability to suggest multiple possible target referent object sets during their turn. If the listener identifies the target object, they can declare ``I know the target object! It is `target object`!''
\\\\
During each turn, the speaker should carefully choose a feature of the target object that delivers the most valuable information to the listener, while the listener adjusts their possible target referent objects based on the previous turn's information.
\\\\
Remember, the listener can only update their referent set from the previous turn's guess; they cannot add new referents.
\\\\
The scoring framework rewards efficiency: the fewer turns required to identify the target object, the higher the score achieved.
\end{tcolorbox}

For Competitive Taboo, the following prompt was used in the implementation:
\begin{tcolorbox}[
    enhanced,
    colback=white, colframe=black!20, 
    boxrule=0.2mm,
    arc=0.5mm, 
    fontupper=\ttfamily\scriptsize,
    fonttitle=\scriptsize\bfseries,
    title=,
    breakable
]
\textbf{\underline{Representative Competitive Taboo Task Prompt:}}\\\\
Play the game of Adversarial Taboo. In this game, there are two players, an attacker and a defender.
\\\\
At the beginning, the attacker is assigned a target word, with which the defender is not informed. The task of the attacker is to induce the defender to utter the target word unconsciously, and then the attacker wins the game. However, the attacker is not allowed to utter the target word; otherwise, the attacker loses.
\\\\
At the same time, the defender tries to figure out the target word. If the defender identifies the target word, he can say ``I know the word! It is `target word`!''. Then the defender wins. Remember, the defender can only guess the word once; otherwise, he will directly lose the game.
\\\\
The game has $\{ \text{max\_turns} \}$ turns. if nobody wins during $\{ \text{max\_turns} \}$ turns, there is a tie.
\end{tcolorbox}

\subsection{Role Prompts}
\paragraph{Cooperative RSA.}
In the task of Cooperative RSA, the prompts for the speaker and the listener are as follows:
\begin{tcolorbox}[
    enhanced,
    colback=white, colframe=black!20, 
    boxrule=0.2mm,
    arc=0.5mm,
    fontupper=\ttfamily\scriptsize,
    fonttitle=\scriptsize\bfseries
]
\textbf{\underline{Speaker Prompt:}}\\\\
\#\#\# Instruction: You are the pragmatic rational speaker. The target object is `$\{ \text{target} \}$' and the object list is `$\{ \text{object\_list} \}$'. Provide your response, including the object feature.
\\\\
\#\#\# Response:
\\\\
\textbf{\underline{Listener Prompt:}}
\\\\
\#\#\# Instruction: You are the pragmatic rational listener. The object list is `$\{ \text{object\_list} \}$'. Provide your inferred target object or the possible target object sets.
\\\\
\#\#\# Response:
\end{tcolorbox}

\paragraph{Competitive Taboo.}
In the task of Competitive Taboo, the prompts for the attacker and the defender are as follows:
\begin{tcolorbox}[
    enhanced,
    colback=white, colframe=black!20, 
    boxrule=0.2mm,
    arc=0.5mm, 
    fontupper=\ttfamily\scriptsize,
    fonttitle=\footnotesize\bfseries
]
\textbf{\underline{Attacker Prompt:}}\\\\
\#\#\# Instruction: You are the attacker. The target word is `{target}`. Provide your response for the next turn.
\\\\
\#\#\# Response:
\\\\
\textbf{\underline{Defender Prompt:}}\\\\
\#\#\# Instruction: You are the defender. Provide your response to infer the target word.
\\\\
\#\#\# Response:
\end{tcolorbox}

\end{document}